\documentclass[aps,a4paper,10pt,twocolumn,footinbib,eqsecnum,showpacs]{revtex4-2} 
\usepackage[T1]{fontenc}
\usepackage{mathptmx}
\usepackage{datetime}
\usepackage{amsmath}
\usepackage{amsfonts}
\usepackage{eufrak}
\usepackage{mathrsfs}
\usepackage[mathscr]{euscript}
\usepackage{mathtools}
\usepackage{textcomp} 
\usepackage{fancyhdr}
\usepackage{colordvi}
\usepackage{hyperref} 
\usepackage{epsfig}
\usepackage{color}
\usepackage{bm}
\usepackage{stmaryrd}
\usepackage{cancel}
\usepackage{tensor}
\usepackage[normalem]{ulem} 
 \advance \textheight by -45mm
\newdateformat{yymmdddate}{\THEYEAR/\twodigit{\THEMONTH}/\twodigit{\THEDAY}}

\def\plotthreewidth{0.7\columnwidth}

\newcommand{\FIGDIR}[1]{#1}

\newcommand{\Dict}[1]{\mathsf{#1}}
\newcommand{\World}[1]{\MakeTextUppercase{#1}}
\newcommand{\Sub}[1]{\MakeTextLowercase{#1}}

\def\DcSize{\World{W}} 
\def\SySize{\World{N}} 
\def\pWordLen{\Omega} 
\def\pFfork{\nu}           

\def\DcSsub{\Sub{W}}
\def\SySsub{\Sub{N}}

\begin{document}

\title{Stylized innovation:\\
 generating timelines by 
 interrogating incrementally available randomised dictionaries}

\author{Paul Kinsler}
\email{Dr.Paul.Kinsler@physics.org}


\affiliation{
  Physics Department,
  Lancaster University,
  Lancaster LA1 4YB,
  United Kingdom.}

\begin{abstract}

A key challenge when trying to understand innovation
 is that it is a dynamic, 
 ongoing process, 
 which can be highly contingent on ephemeral factors
 such as culture, 
 economics, 
 or luck.
This means that
 any analysis of the real-world process must necessarily be historical --
 and thus probably too late to be most useful -- 
 but also cannot be sure what the properties 
 of the web of connections 
 between innovations is or was.
Here I try to address this by designing and generating a set of
 synthetic innovation web ``dictionaries''
 that can be used to host sampled innovation timelines, 
 probe the overall statistics and behaviours of these processes, 
 and determine the degree of their reliance on the 
 structure or generating algorithm.
Thus, 
 inspired by the work of 
 Fink, Reeves, Palma and Farr (2016) 
 on innovation in language, gastronomy, and technology,
 I study how new symbol discovery 
 manifests itself in terms of additional ``word'' vocabulary 
 being available from dictionaries generated from a finite number of symbols.
Several distinct dictionary generation models are investigated
 using numerical simulation,
 with emphasis on the scaling of knowledge
 as dictionary generators and parameters
 are varied, 
 and the role of which order
 the symbols are discovered in.

\end{abstract}

\lhead{\includegraphics[height=5mm,angle=0]{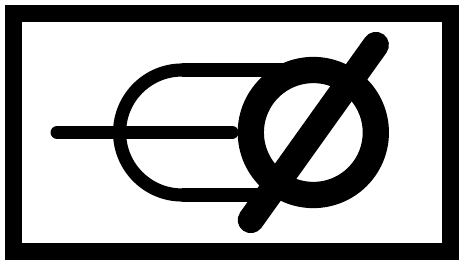}~~INNOBET}
\chead{Stylized Innovation in Synthetic Dictionaries}
\rhead{
\href{mailto:Dr.Paul.Kinsler@physics.org}{Dr.Paul.Kinsler@physics.org}\\
\href{http://www.kinsler.org/physics/}{http://www.kinsler.org/physics/}
}

\date{\today}
\maketitle
\thispagestyle{fancy}

\section{Introduction}
\label{S-Introduction}

Fink et al. \cite{Fink-RPF-2016arxiv}
 recently presented an interesting discussion and analysis of 
 serendipity and strategy in innovation scenarios, 
 paying particular attention to real-world datasets 
 based on 
 language, 
 gastronomy, 
 and technology.
Here, 
 however, 
 I take a look at the same type of innovation/discovery processes
 in the context of synthetic datasets
 where I have designed a range
 of different underlying ``innovation'' spaces to investigate.
Although this is clearly not as realistic 
 as using real-world datasets, 
 it allows us to 
 (a) try to establish both individual and universal features
 of a discovery/innovation process, 
 and 
 (b) avoids us having to make ad hoc
 judgments about
 whether chosen real items are individual components or compounds
 of other components.

In this paper I will investigate the synthetic datasets
 in a way broadly compatible
 with Fink et al. \cite{Fink-RPF-2016arxiv}, 
 but emphasizing discovery over determination of global properties, 
 and relying primarily on numerical calculations.
Notably, 
 I will use some specific terminology that differs from theirs; 
 this is in order to avoid unwanted inferences 
 by clarifying the basic abstractions underpinning
 the dataset generation and its analysis.
The total dataset will be called a \emph{dictionary}, 
 or world-dictionary,
 which is a list compiled of \emph{words}.
Each word in the dictionary is 
 a string of \emph{symbols} that has been
 randomly generated by an algorithm.
As we \emph{discover} extra symbols
 from the allowed list, 
 we can \emph{interrogate} the dictionary, 
 finding ``innovative'' new words to add to our
 ``sub-dictionary'' of known words.

We will see how the process of ever-better interrogation of the dictionaries
 plays out in average terms and
 in some specific examples.
Two different symbols discovery mechanisms are of interest here --
 most common first (i.e. frequency-order), 
 and randomised.
Of course,
 in many contexts we might think it most likely 
 that the next-discovered symbol will be the most frequently occurring.
Alternatively, 
 when chasing innovation based on assembling known components, 
 as might be true of cooking recipes or software systems, 
 we can define an average usefulness \cite{Fink-RPF-2016arxiv}
 on which to base an intelligent symbol-acquisition strategy.

However, 
 when interrogating unknown systems, 
 we may need to just rely for innovation on whatever new thing we have 
 managed to discover.
This will be 
 without knowing in advance of use
 how frequently it will turn out to be useful --
 because we do not have access to the world-dictionary of our unknown system.
\emph{This ``explorative'' (or perhaps even ``evolutionary'') view of innovation
 is distinct from that of the more ``assembly''-focused view
 of combining known things}.
Therefore I also show results for discovering the symbols in a random-order, 
 where indeed we {can} see the sudden bursts of discovery
 we might normally expect from terms like ``innovation'' and ``serendipity''.
Further, 
 although the combination view of innovation might be most easily
 related to the idea of ``assembly theory'' as used 
 to describe molecular assembly trees
 \cite{Marshall-MC-2017rsptam,Liu-MBMWC-2021sa}, 
 one could imagine recasting that in the context
 of my dictionary interrogation approach; 
 or indeed using algorithms -- 
 such the various dictionary generation algorithms I propose here -- 
 as a testbed for assembly theory in a more abstract or general context.


Unlike the treatment in Sood et al \cite{Sood-MSGP-2010prl}, 
 my world dictionaries are static,
 and do not involve components (symbols) 
 which appear as new, 
 and then later may die off and disappear, 
 taking the words they appear in with them.
However, 
 one might -- 
 although I do not do so here --
 consider interrogations where symbols are not only learnt 
 but forgotten, 
 so that the currently known sub-dictionary of the world-dictionary
 will never (or is unlikely to) 
 reach a state of total knowledge.
Instead, 
 the subset of known symbols would only ever provide
 a sort of moving window on to the world-dictionary.



Other distinct models worthy of note here 
 are the agent-decision led, 
 persuasion/contagion-like models 
 of Weiss et al \cite{Weiss-PGPPBWN-2014prx}, 
 and the component-optimization view
 taken by McNerney et al \cite{McNerny-FRT-2011pnas}
 which led to nice power-law behaviour.
However, 
 as implied above, 
 my model here follows more along the lines of 
 ``the system (dictionary) will only let me
 create a finite number of compounds (words),
   but I don't know what the allowed (symbols and words) are,
   and will only be able to find out as I discover new things (symbols)''.

In what follows I will present and discuss 
 the result of some numerical experiments
 based on algorithmically constructed dictionaries.
Although real-world ``dictionaries'' are available, 
 their underlying structure --
 the equivalent of the generating algorithm I use here 
 to create my synthetic dictionaries --
 is unknown (or may not exist in any meaningful sense), 
 which hampers analysis.
In Sec. \ref{S-Dictionaries}
 I will present the generating algorithms
 along with the averaged scaling behaviour of discovery processes.
However, 
 since and real ``innovation'' process 
 will only experience one possible symbol discovery order, 
 in Sec. \ref{S-Discovery}
 I present and comment on specific discovery processes, 
 noting their main features 
 and discussing and comparing the outcomes.
Finally, 
 in Sec. \ref{S-Conclusion}, 
 I present my conclusions.

\section{Dictionaries and Scaling}
\label{S-Dictionaries}

Here the dictionaries I investigate are compiled of words
 made from strings of symbols generated by 
 a randomly driven algorithm.
However, 
 we might consider the symbols to represent ingredients
 to be assembled into recipes (``words''), 
 and then the compendium of recipes make up a recipe-book (``dictionary''); 
 or they might be gestures and movements which we assemble into dance moves, 
 and then into a whole dance routine.
At any point in the discovery process, 
 we will know only a subset of all the allowed symbols, 
 and hence only have found (``innovated'') a subset of the words; 
 i.e. we will be working from not the total dictionary, 
 but a sub-dictionary.

Here I consider how new words (``innovations'')
 become available 
 as new symbols are found and used to to interrogate the dictionary, 
 and thus allow us to expand our sub-dictionary.
For this we rely on aggregate quantities
 made up from all changes occurring during the process of discovering all symbols
 and hence the whole dictionary.

Fink et al \cite{Fink-RPF-2016arxiv} used an elegant series of diagrams
 that ranked symbols (their `components')
 based on their ``average usefulness $\bar{u}_\alpha(n)$'', 
 where the (ordinary) usefulness ${u}_\alpha$ of symbol $\alpha$
 was a simple numerical count
 of how many words (their `products') was present in.
Here I prefer to focus on the usefulness as might be estimated
 due to the current state of knowledge,  
 rather than an average usefulness  \cite{Fink-RPF-2016arxiv}
 which involves a sum over all the alternative selections of other symbols.
This less comprehensive averaging means we expect to see less stable rankings
 as the discovery process proceeds than Fink et al would, 
 but it does not presume a knowledge of the (world) dictionary.
Further, 
 here I am most interested in the variation in the rates of change, 
 not any absolute value.

The measures I consider when investigating the scaling of innovation are
 all normalized so that according to some simple assumptions, 
 the full symbol discovery process on some dictionary
 will return that innovation measure to 
 depend on the size of its symbol list, 
 i.e. be $\SySize-1$.
However, 
 since the simple assumptions are in fact unlikely to hold in any given case, 
 the difference between the calculated value and the nominal `assumed value'
 will provide information about the innovation process.
The measures are:

\begin{enumerate}

\item
 \emph{Changes in ranks}
 $\delta_r$, 
  a count of how many symbol-usefulness rankings changed
  at every new discovery,
  scaled down by $\SySsub$, 
  i.e. the number of known symbols $\SySsub$.
 This is chosen so that if every known symbol changes rank
  at every new discovery, 
  for each of the $\SySize-1$ discoveries,
  then $\delta_r=\SySize-1$.

\item
 \emph{Linear ranks shift}
 $\delta_\omega$,
  a sum of all the absolute values of changes in symbol-usefulness
  at every new discovery,
  scaled down by $\SySsub^2/2$.  
 This is chosen so that if every known symbol changes usefulness
  by an average $\SySsub/2$
  at every new discovery, 
  then $\delta_\omega=\SySize-1$.

\item
 \emph{Nonlinear ranks shift}
 $\delta_\chi$,
  a sum of all the changes in symbol-usefulness \emph{squared}
  at every new discovery,
  scaled by $\SySsub^3/4$.
 This is chosen so that if every known symbol changes usefulness
  by an average $\SySsub/2$
  at every new discovery, 
  then $\delta_\chi=\SySize-1$;
  with the squaring acting
  to de-emphasize the smaller changes.

\end{enumerate}

I now summarize and describe the dictionary generation methods I tested, 
 and how the aggregate quantities varied with symbol count $\SySize$, 
 dictionary size $\DcSize$, 
 and any generator-specific parameters.
Each result in this section is an average
 taken from an ensemble of at least $16$ test dictionaries,
 with more test dictionaries generated if necessary
 so as to keep the relative standard deviation
 in the aggregate quantities below $5\%$.
In the rest of this section 
 all data plotted is summed over the entire discovery process, 
 from first symbol to last.

\subsection{Fake random dictionaries: ``null'' or  N-dictionaries}

These ``null'' dictionary are not real dictionaries --  
 no word list exists, 
 just a nominal word count of $\DcSize$
 and a nominal number of symbols $\SySize$.
They are defined solely as an algorithm 
 intended to test the
 normalization of the $\delta$ quantities, 
 and to provide a baseline of comparison that is unaffected
 by any specifics of a dictionary generation algorithm.
If such an N-dictionary is interrogated, 
 it returns information that the 
 dictionary size is scaled in proportion to the number of known symbols
 (i.e. that $\DcSsub = \SySsub \DcSize / \SySize$,
 and that the allowed symbols
 have randomly ordered frequencies.
Note that any re-interrogation of an N-dictionary
 will return a different randomised set of frequencies.

Fig. \ref{fig-null-scaling} compares the innovation measures 
 for two symbol discovery possibilities:
 (a) discovery in order of frequency (or usefulness)
 in the world dictionary, 
 or, 
 (d) discovery in random-order.
The two sets of results here 
 span a range of symbol list and dictionary sizes.
We see that for randomly chosen discovery order, 
 $\delta_r \sim \SySize$ within random error, as expected.
The values for $\delta_\omega$ and $\delta_\chi$
 are somewhat suppressed below the $\SySize$ value, 
 as expected since it is very unlikely
 that any two subsequent interrogations will be 
 as different as assumed by the normalization.
As should be expected, 
 the two discovery orders give results that differ only due
 to sampling error, 
 since the ``frequency-order'' is randomly re-set by the algorithm 
 at each subdictionary interrogation.

\def\STANDARDCAP{Here, 
 frequency-order discovery (open mesh) and random-order discovery (colour)
 results for ~Changes in ranks $\delta_r$, 
 ~Linear shifts in ranks $\delta_\omega$, 
 and ~Nonlinear shifts in ranks $\delta_\chi$.
Since the open mesh is often hard to see easily, 
 the colour-coding of the random-order result surface
 is done according to the difference between the two sets of data.
}

\begin{figure}[h]
\includegraphics[angle=0,width=\plotthreewidth]{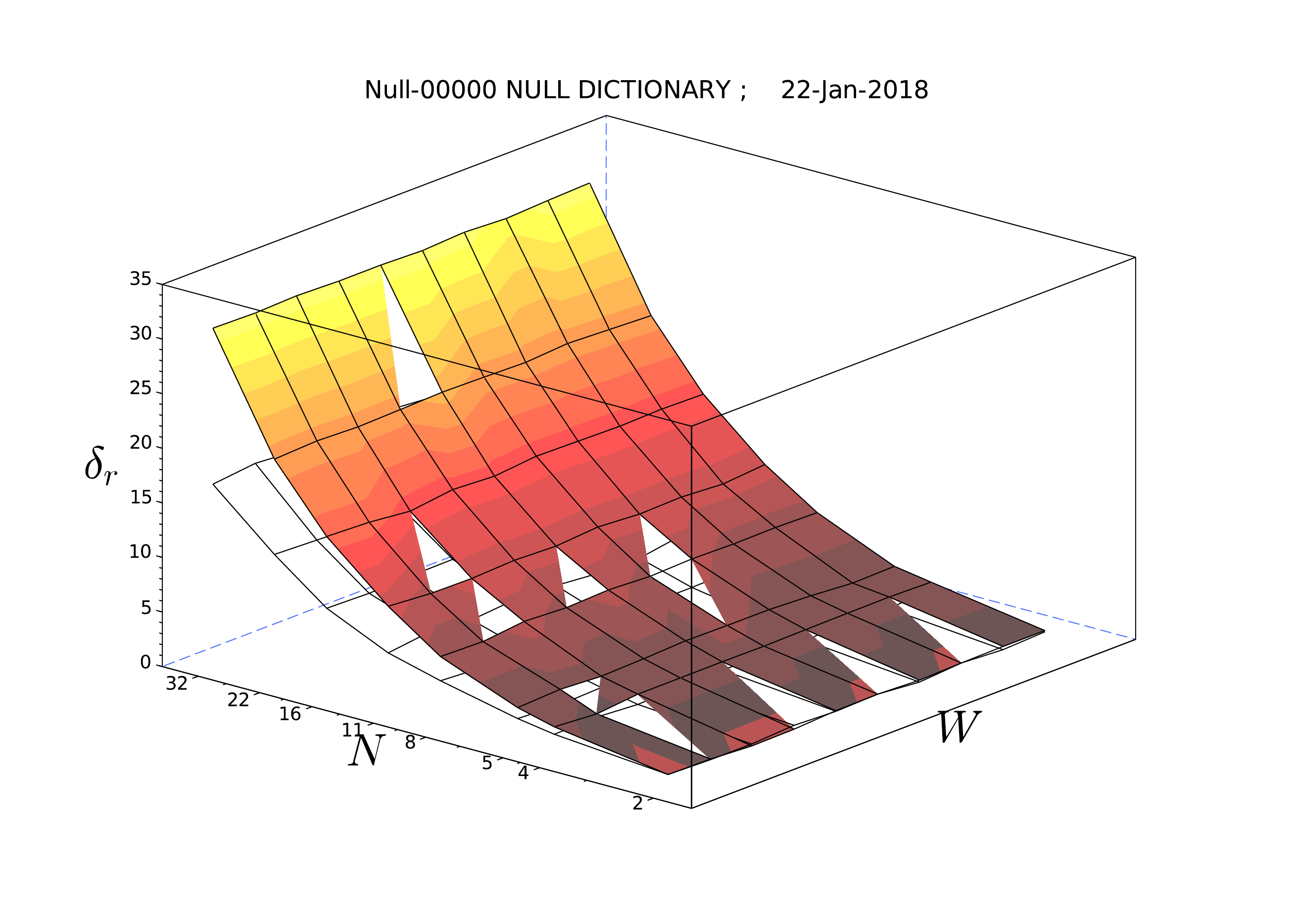}
\includegraphics[angle=0,width=\plotthreewidth]{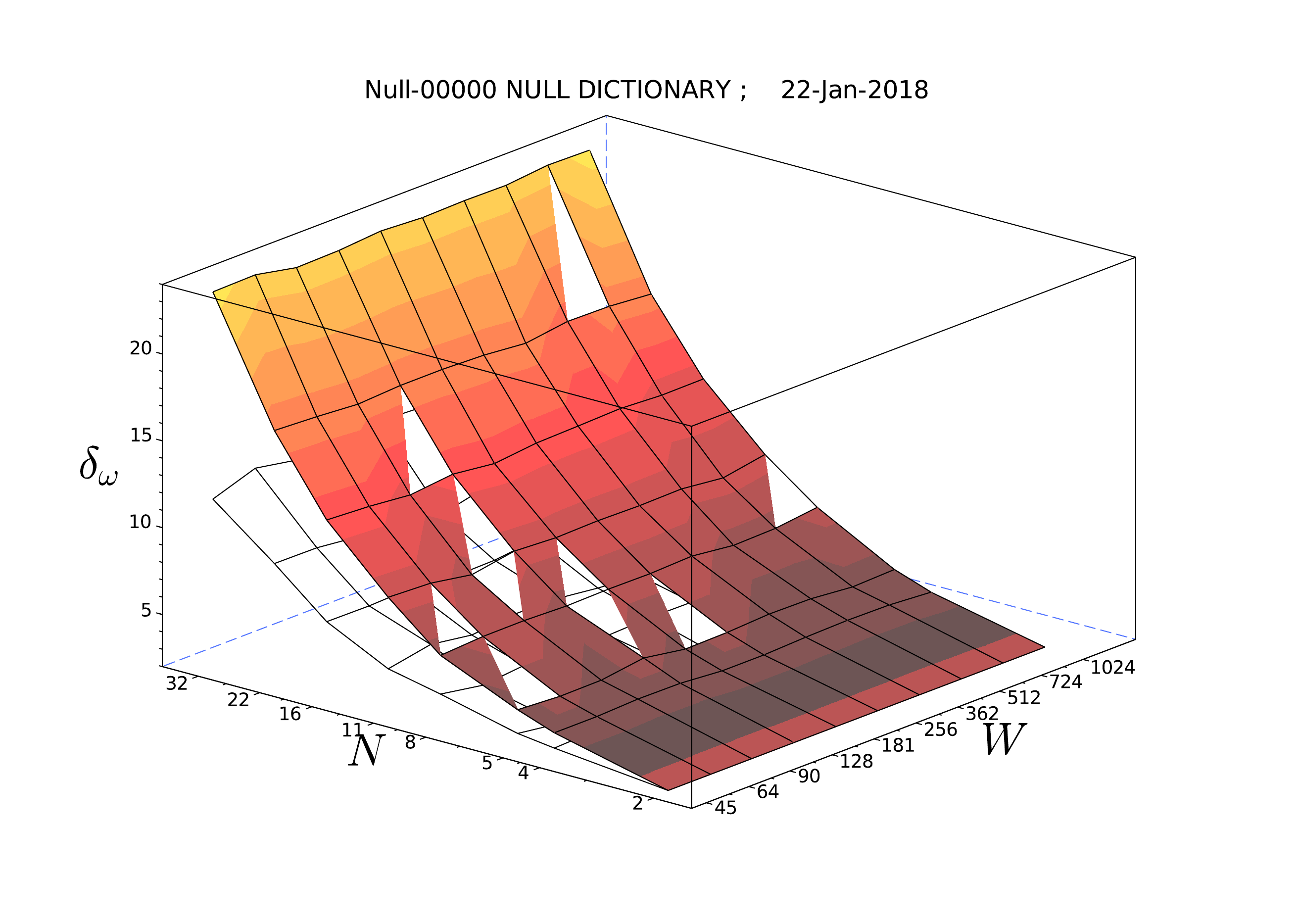}
\includegraphics[angle=0,width=\plotthreewidth]{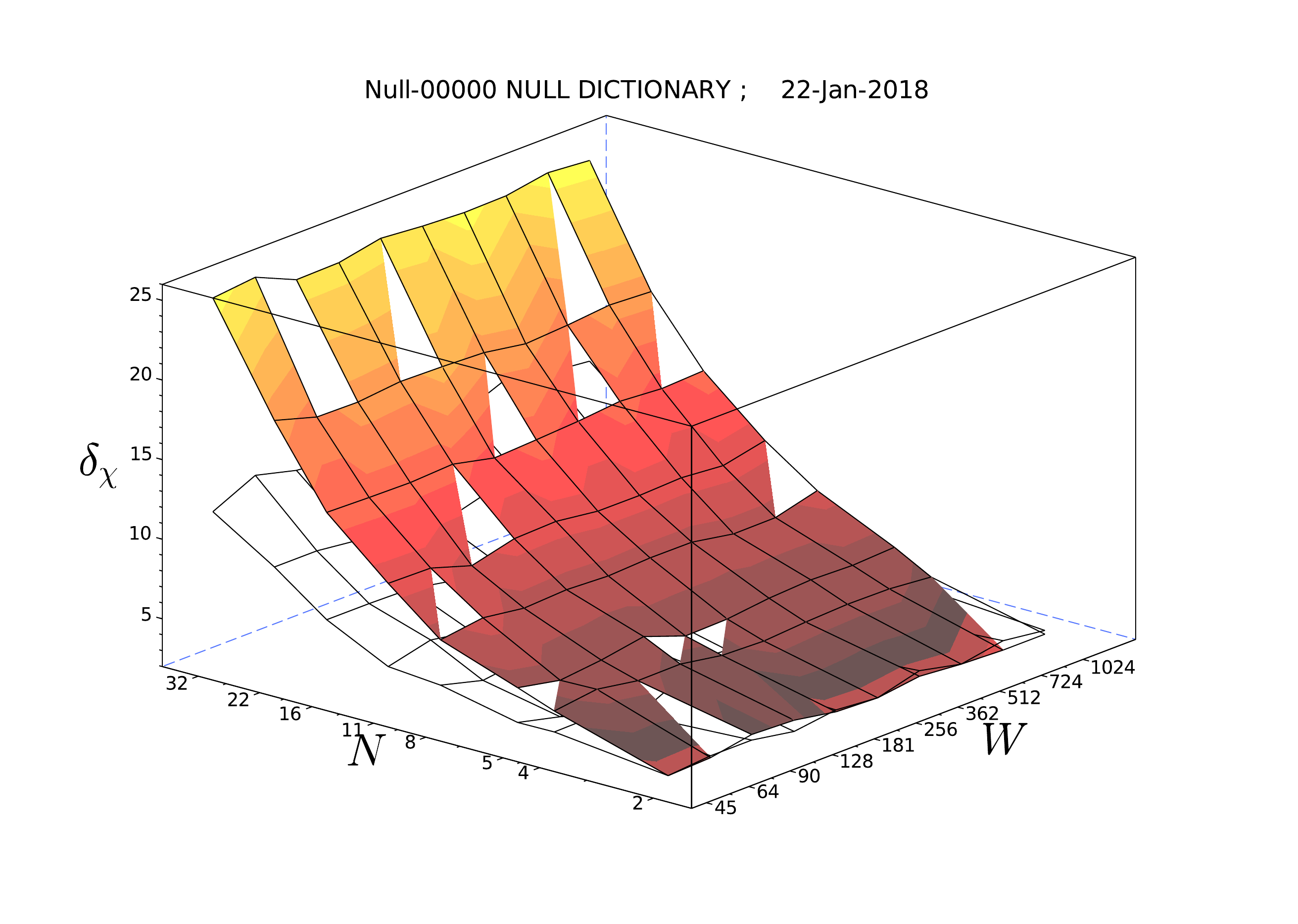}
\caption{Scaling of innovation in
 $\Dict{N}^{\DcSize}_{\SySize}$ null random dictionaries, 
 with symbol number $\SySize$ varying on a log scale from $2$ to $32$, 
 and dictionary size $\DcSize$ on a log scale from $45$ to $1024$. \\
\STANDARDCAP
}
\label{fig-null-scaling}
\end{figure}


\subsection{Fixed length words: F-dictionaries}

A fixed length word dictionary $\Dict{F}^{\DcSize}_{\SySize,\pWordLen}$
 containing $\DcSize$ words 
 can be straightforwardly generated
 using a symbol list of length $\SySize$, 
 with $\pWordLen$ symbols per word, 
 each symbol being chosen randomly.
Words can contain any number (up to $\pWordLen$) 
 of repeated symbols, 
 and words may be repeated in the dictionary.

With this generator, 
 we will want ${\DcSize} \ll {\SySize}^{\pWordLen}$ 
 so that the random sampling of all possible words
 plays a strong role when populating the dictionary.

\begin{figure}[h]
\includegraphics[angle=0,width=\plotthreewidth]{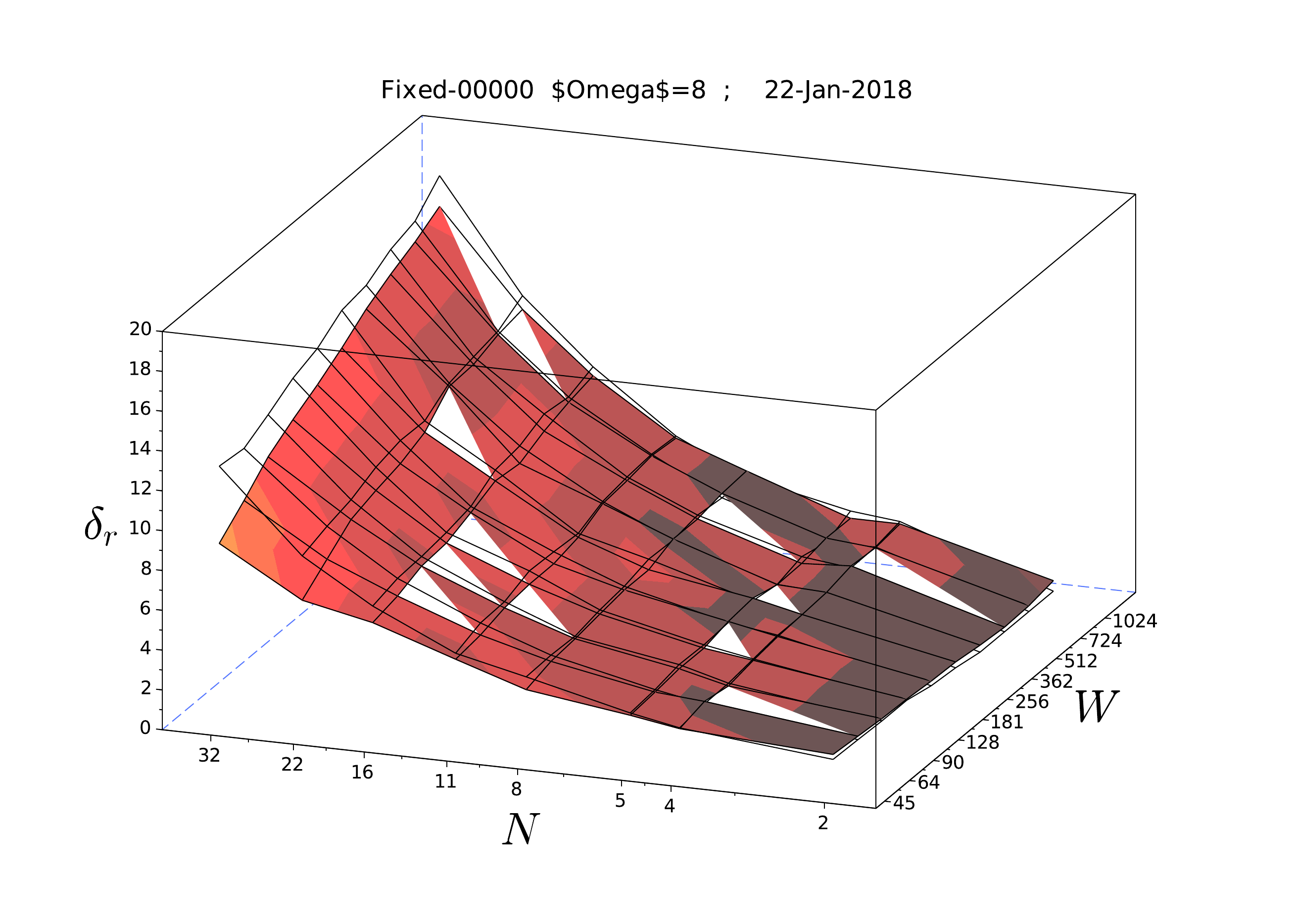}
\includegraphics[angle=0,width=\plotthreewidth]{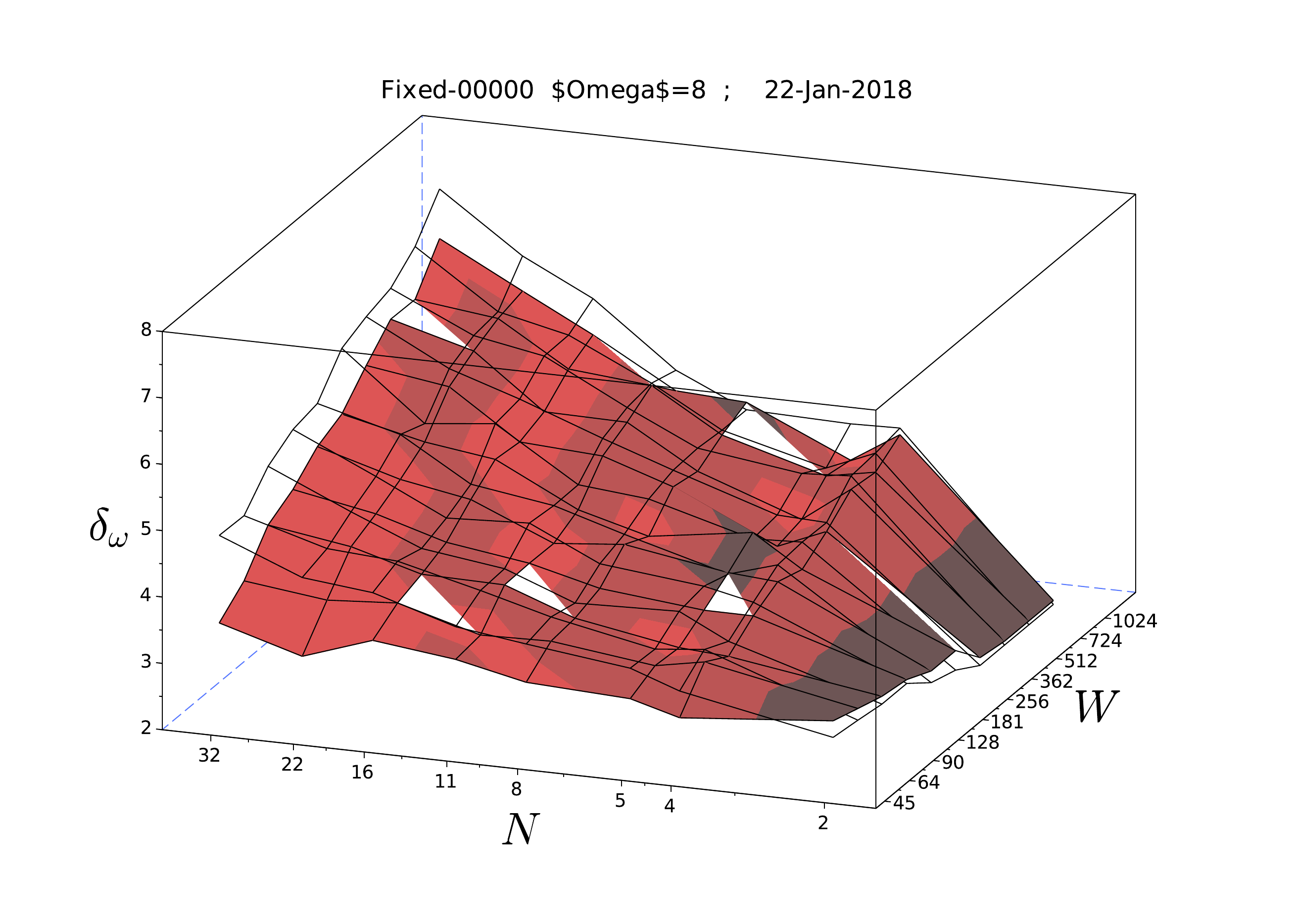}
\includegraphics[angle=0,width=\plotthreewidth]{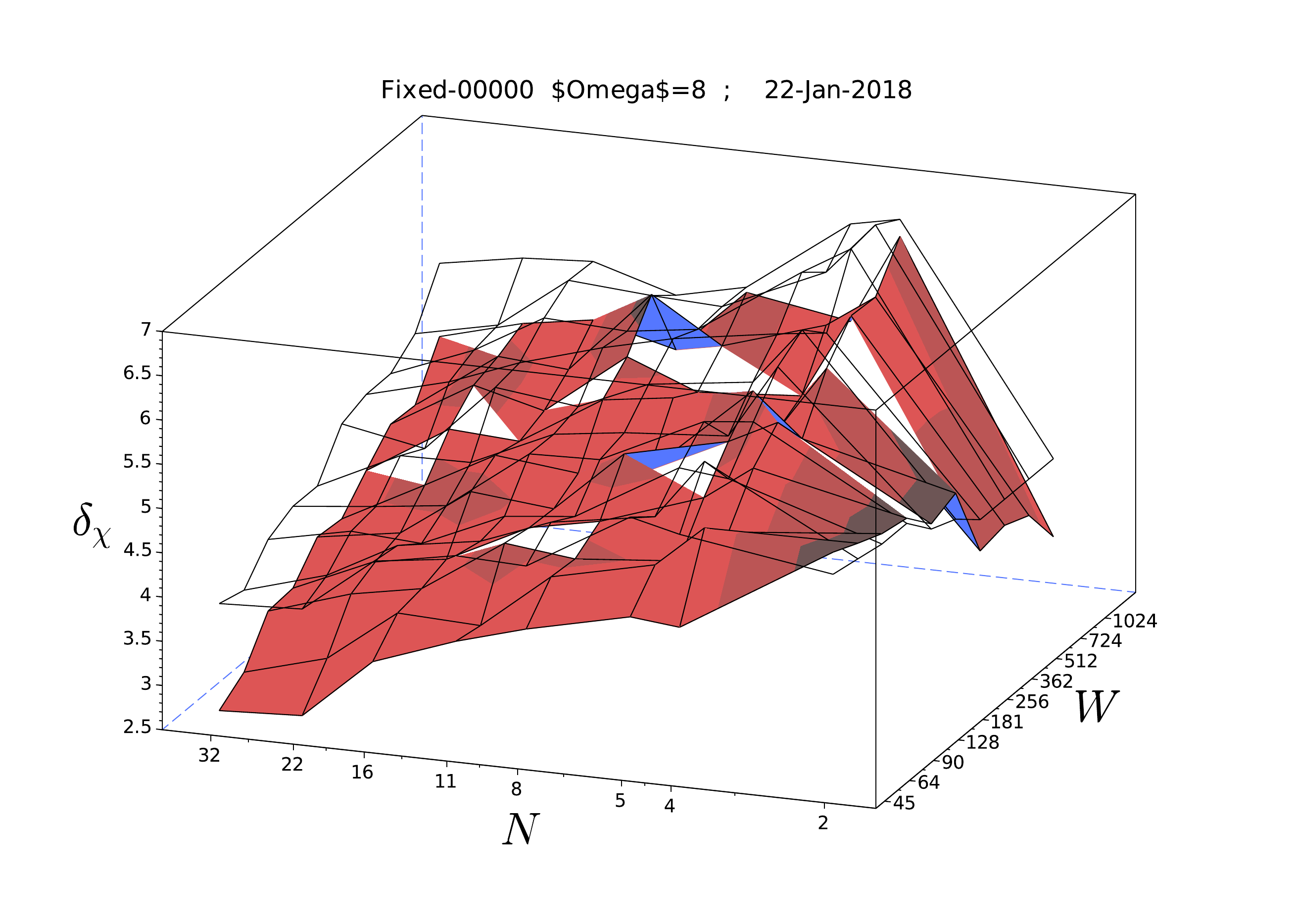}
\caption{Scaling of innovation in
 $\Dict{F}^{\DcSize}_{\SySize,8}$ fixed random dictionaries, 
 with symbol number $\SySize$ varying on a log scale from $2$ to $32$, 
 and dictionary size $\DcSize$ on a log scale from $45$ to $1024$. \\
\STANDARDCAP
}
\label{fig-fixed-scaling}
\end{figure}

\begin{figure}[h]
\includegraphics[angle=0,width=\plotthreewidth]{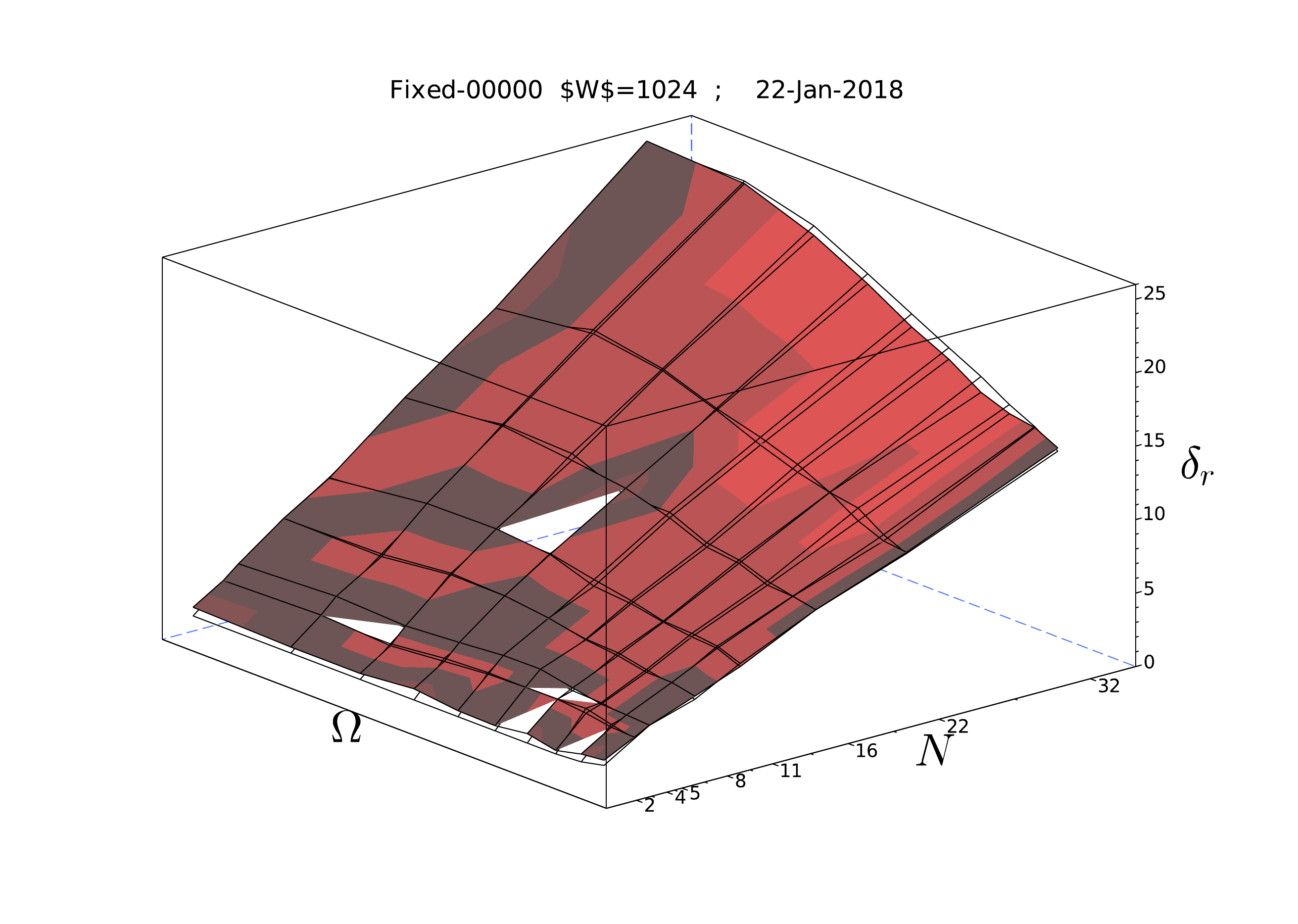}
\includegraphics[angle=0,width=\plotthreewidth]{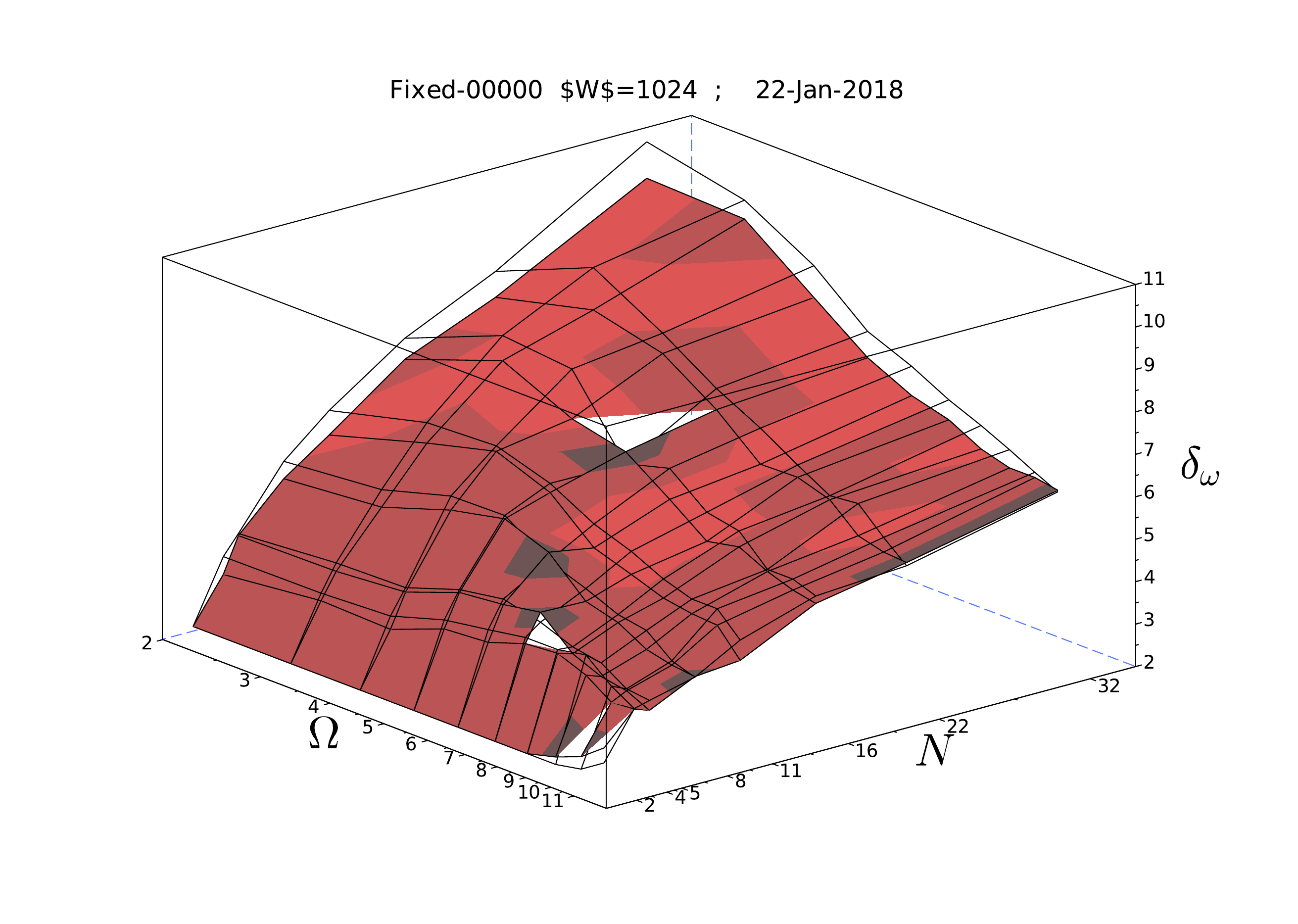}
\includegraphics[angle=0,width=\plotthreewidth]{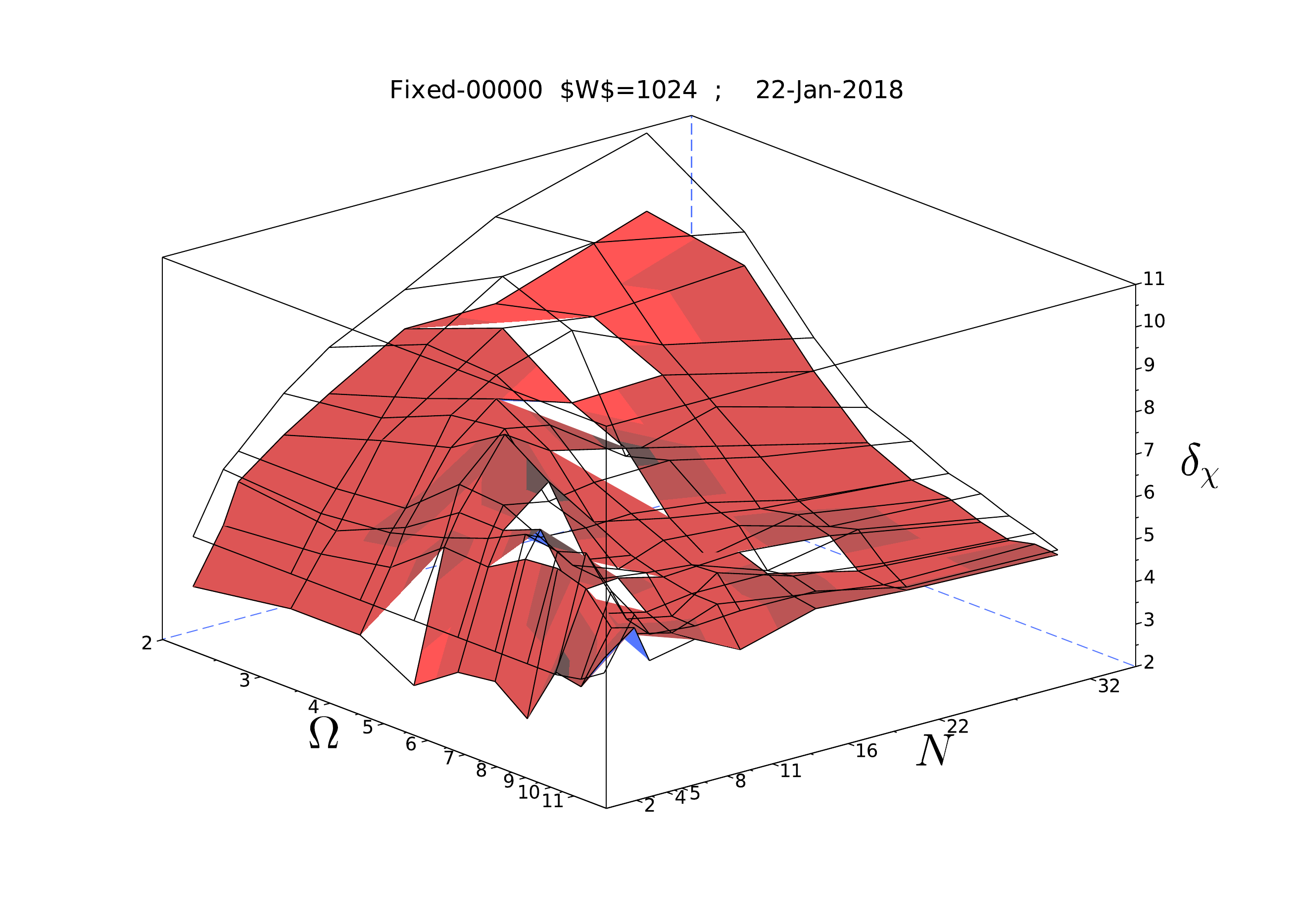}
\caption{Scaling of innovation in
 $\Dict{F}^{1024}_{\SySize,\pWordLen}$ fixed random dictionaries, 
 with symbol number $\SySize$ varying on a log scale from $2$ to $32$, 
 and word length $\pWordLen$ from $2$ to $11$. \\
\STANDARDCAP
}
\label{fig-fixed-parscaling}
\end{figure}

Fig. \ref{fig-fixed-scaling} compares the innovation measures 
 for two symbol discovery possibilities:
 (a) discovery the frequency-order of symbols in the world dictionary, 
 or, 
 (b) discovery in random-order.
The two sets of results here 
 span a range of symbol list and dictionary sizes, 
 but are constrained by a fixed word length $\pWordLen=8$.
Since the dictionary parameters do always not honour
 the sampling inequality above, 
 particularly for the smallest symbol lists, 
 we see a pronounced `knee' in each of the $\delta$ measures
 below $\SySize=4$. 

For large symbol lists and small dictionary sizes
 we see that the $\delta$ measures 
 report higher innovation for frequency-order discovery, 
 although the difference is not large.
This occurs because as the most frequent symbols 
 occurs in the largest fraction of words, 
 and so for any comparable shift in rank or frequency, 
 the measures undergo a lesser rescaling
 due to the normalization process, 
 since $\SySsub$ is smaller earlier in the discovery process.

Perhaps most interestingly, 
 we see that the change count index $\delta_r$
 reports higher innovation for larger symbol lists, 
 in a way approximately with its definition,
 but this behaviour is suppressed for $\delta_\omega$, 
 and even reversed for the nonlinear measure $\delta_\chi$.
This implies that the discovery process is more incremental for 
 larger $\SySize$, 
 involving relatively few(er) discoveries that substantially 
 revolutionize the size and structure
 of the currently known sub-dictionary.
This is not unexpected, 
 since in F-dictionaries with larger word counts, 
 the whole set of symbols will be better and more evenly sampled.

Since the above results were calculate for a fixed word length, 
 in fig. \ref{fig-fixed-parscaling} 
 I instead hold the dictionary size $\DcSize$ fixed at $1024$, 
 and vary word length $\pWordLen$.
There is a clear trend for higher innovation for shorter words, 
 and the trend for suppressed innovations $\delta_\omega$ and $\delta_\chi$
 at larger $\SySize$ systematically fades away for short words, 
 and even reverses for the shortest word lengths.
This is most likely caused by the same reason there was a knee 
 for small dictionary sizes in fig. \ref{fig-fixed-scaling}:
 e.g. taking opposite corners of the parameter ranges, 
 note that with $\SySize=32$ and $\pWordLen=2$, 
 or for $\SySize=2$ and $\pWordLen=10$, 
 the number of unique words is $32^{2}$ or $2^{10}$ --
 both being the same as $\DcSize = 1024$.

\subsection{Extensible: E-dictionaries}

An extensible dictionary $\Dict{E}^{\DcSize}_{\SySize}$
 containing $\DcSize$ words is generated
 using a symbol list of length $\SySize$.
Each new word is added iteratively, 
 starting from a single word
 consisting only of the initial symbol $\alpha_\circ$, 
 randomly chosen new symbols are appended until
 the result is not already in the dictionary.
There is no maximum word length
 and no generator-specific parameters.

\begin{figure}[h]
\includegraphics[angle=0,width=\plotthreewidth]{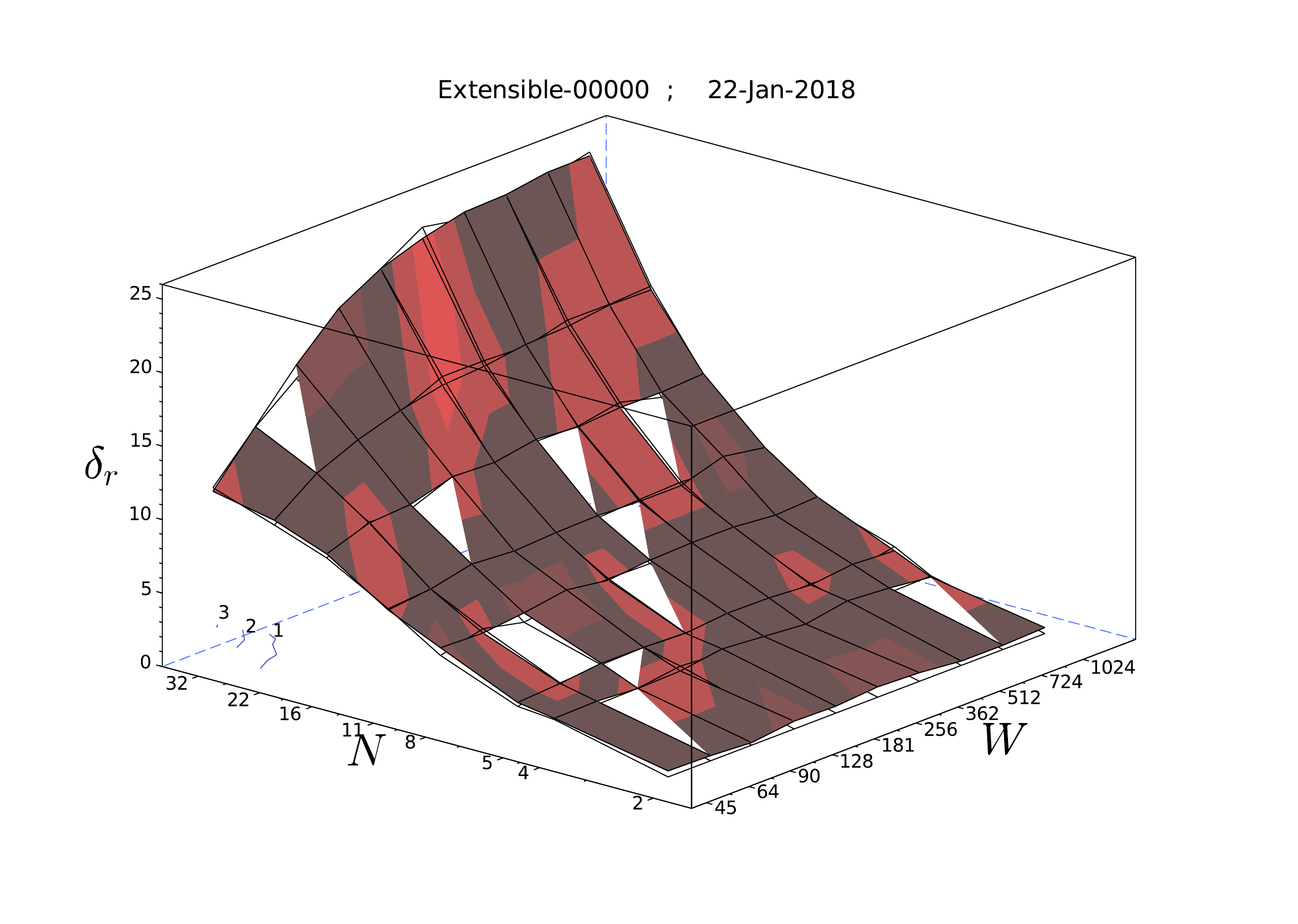}
\includegraphics[angle=0,width=\plotthreewidth]{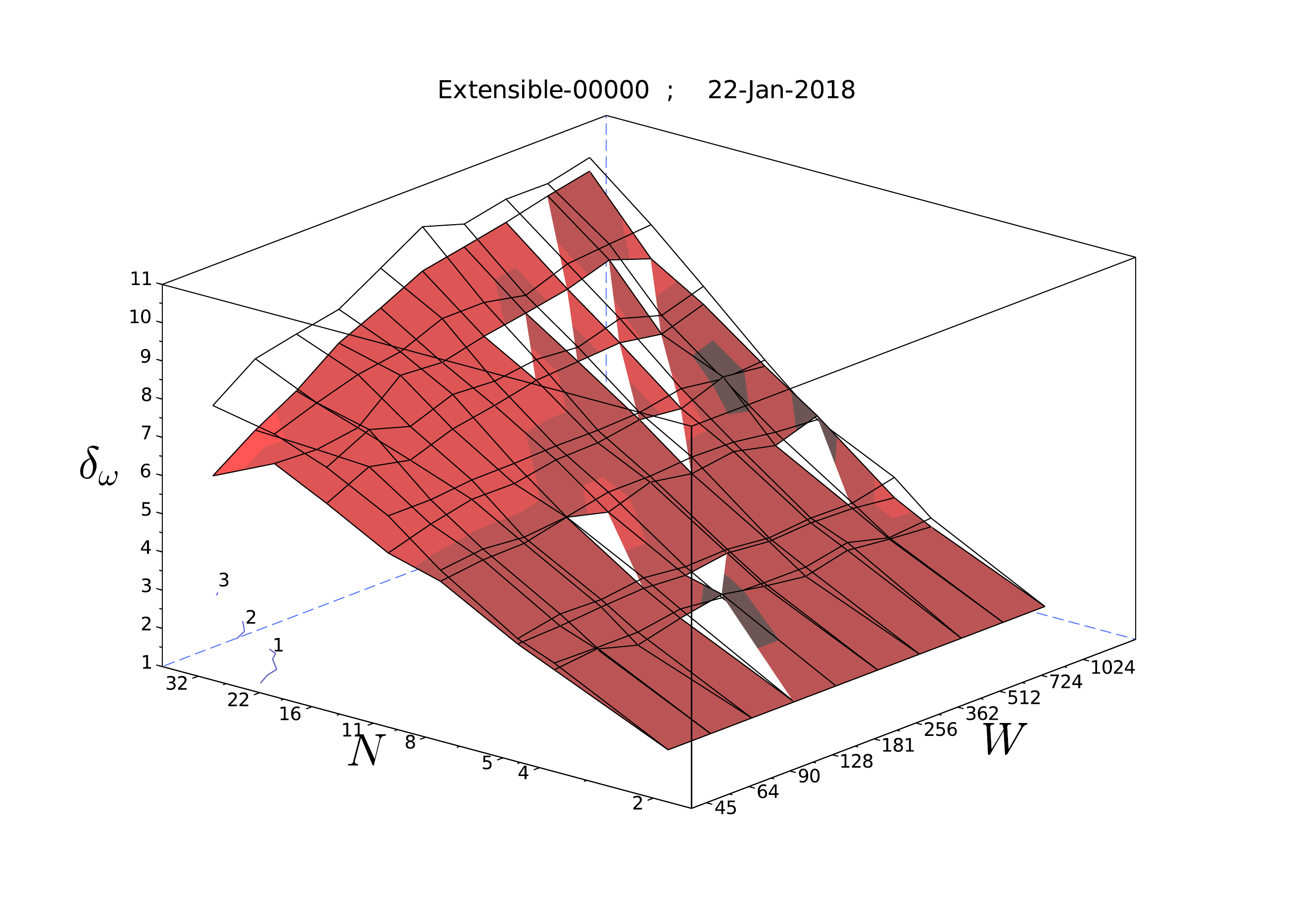}
\includegraphics[angle=0,width=\plotthreewidth]{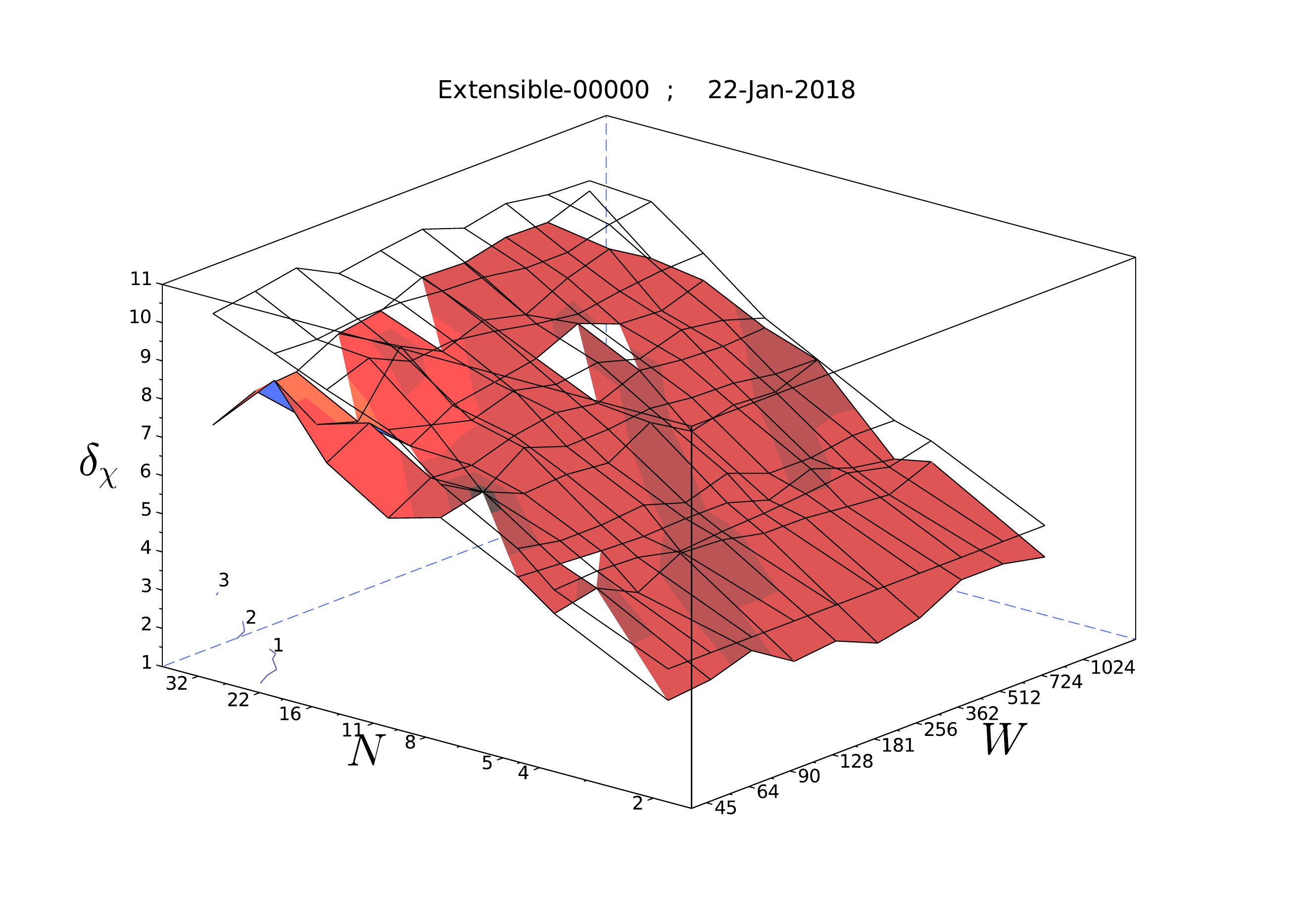}
\caption{Scaling of innovation in
 $\Dict{E}^{\DcSize}_{\SySize}$ extensible dictionaries, 
 with symbol number $\SySize$ varying on a log scale from $2$ to $32$, 
 and dictionary size $\DcSize$ on a log scale from $45$ to $1024$. \\
\STANDARDCAP
}
\label{fig-extensible-scaling}
\end{figure}

Fig. \ref{fig-extensible-scaling} compares the innovation measures 
 for two symbol discovery possibilities:
 (a) discovery the frequency-order of symbols in the world dictionary, 
 or, 
 (d) discovery in random-order.
The two sets of results here 
 span a range of symbol list and dictionary sizes.
For low word counts, 
 and large enough symbol lists, 
 some symbols might not be chosen for use in 
 words in the dictionary; 
 the average number of unused symbols 
 is indicated by the floating contours on the figure.

For large symbol lists and small dictionary sizes
 we see that the $\delta$ innovation measures 
 report higher innovation for frequency-order discovery, 
 although the difference is not large.
The likely reason is the same as for F-dictionaries, 
 and due to the weaker effect of the normalization on early innovative bursts.

Despite
 the very different dictionary generation algorithm, 
 the trends in $\delta_r$ are rather similar to those seen in F-dictionaries, 
 although the measure is more strongly depressed for small dictionaries
 with large symbol lists.
However, 
 the linear and nonlinear measures $\delta_\omega, \delta_\chi$
 behave very differently, 
 being in line with the `expected' behaviour 
 of increasing with symbol list size.

\subsection{Chain: C-dictionaries}

A chain dictionary $\Dict{C}^{\DcSize}_{\SySize,\pFfork}$
 containing $\DcSize$ words is generated
 using a symbol list of length $\SySize$.
It is an incremental process,
 based on a starting point consisting of an initial random symbol
 $\alpha_\circ$.
Each new word is added iteratively:
 either by
 (i) with probability $1-\pFfork$, 
  an existing word being chosen and a new random symbol appended
  to create the new word,
  which is only kept if the result is not already present;
 or 
 (ii) with probability $\pFfork$, 
  another (single) random symbol is chosen to be the new word, 
  which is only kept if the result is not already present.
In this way the ``fork probability'' $\pFfork$ determines how fast 
 the dictionary accumulates words with different starting symbols.
Although we should note that the starting symbol has no special role,
 a low $\pFfork$ means that the early-chosen symbols, 
 and in particular $\alpha_\circ$,
 will dominate 
 the symbol frequencies in the resulting dictionary.

For low word counts, 
 and large enough symbol lists, 
 some symbols might not be chosen for use in 
 words in the dictionary.

\begin{figure}[h]
\includegraphics[angle=0,width=\plotthreewidth]{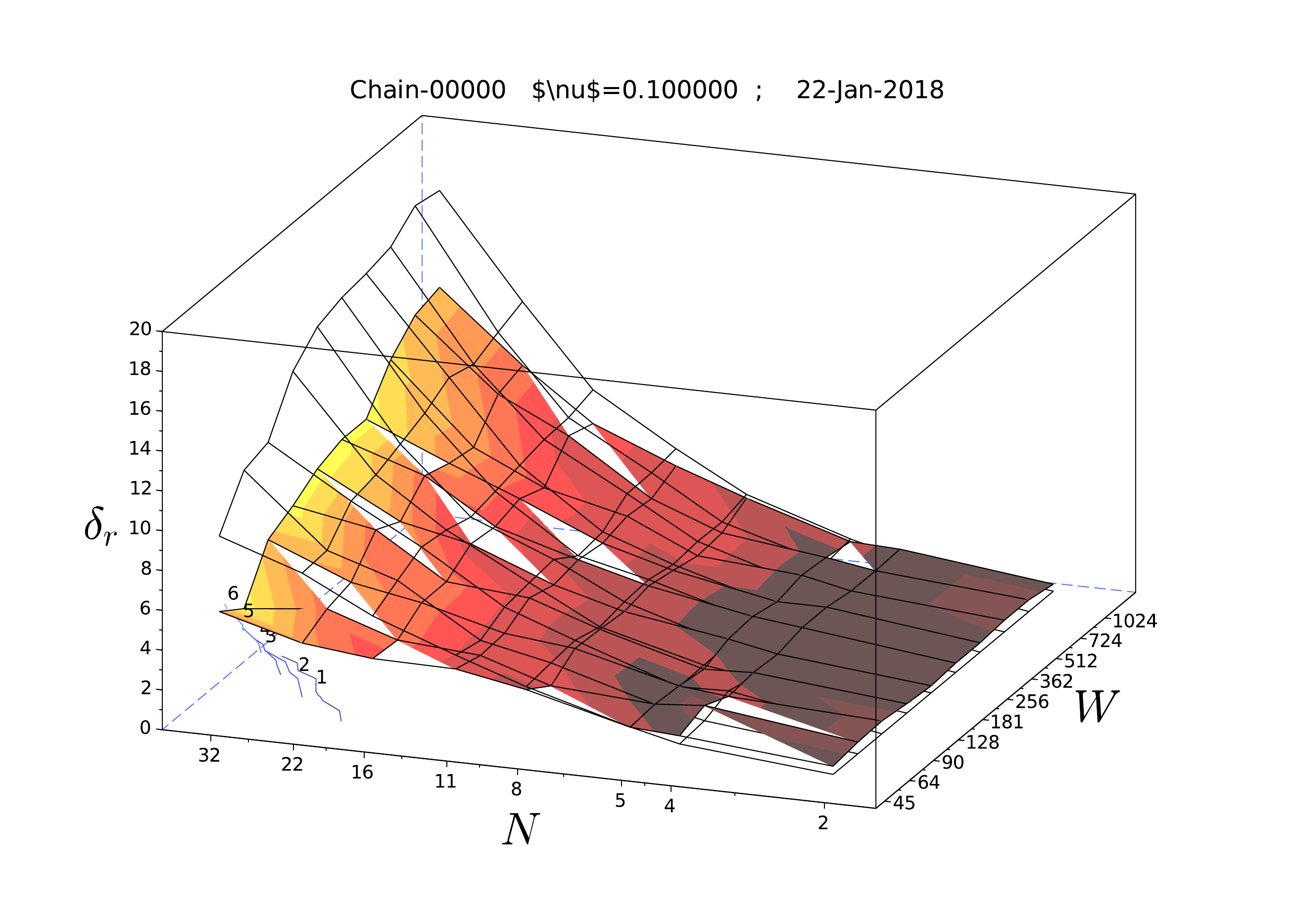}
\includegraphics[angle=0,width=\plotthreewidth]{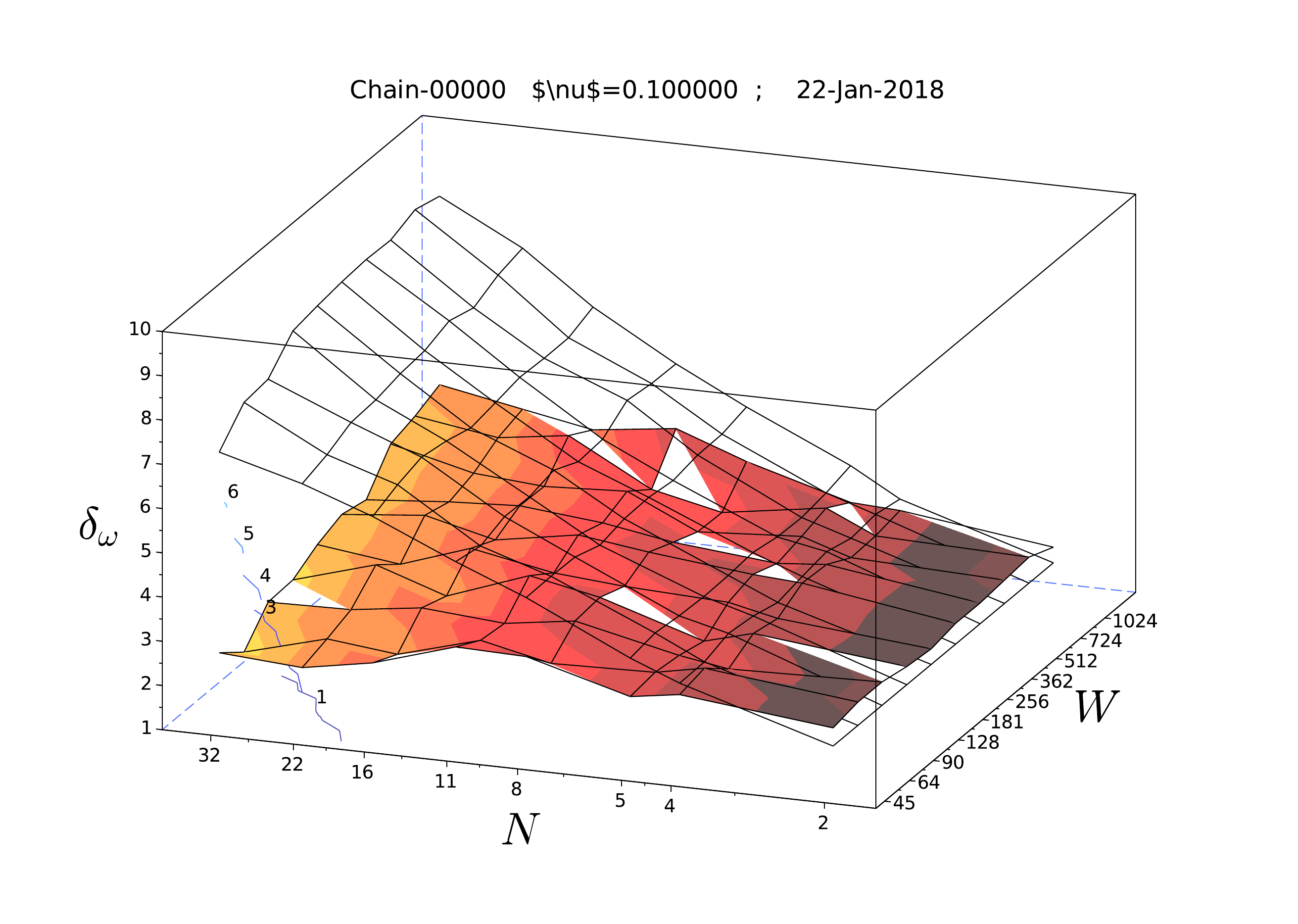}
\includegraphics[angle=0,width=\plotthreewidth]{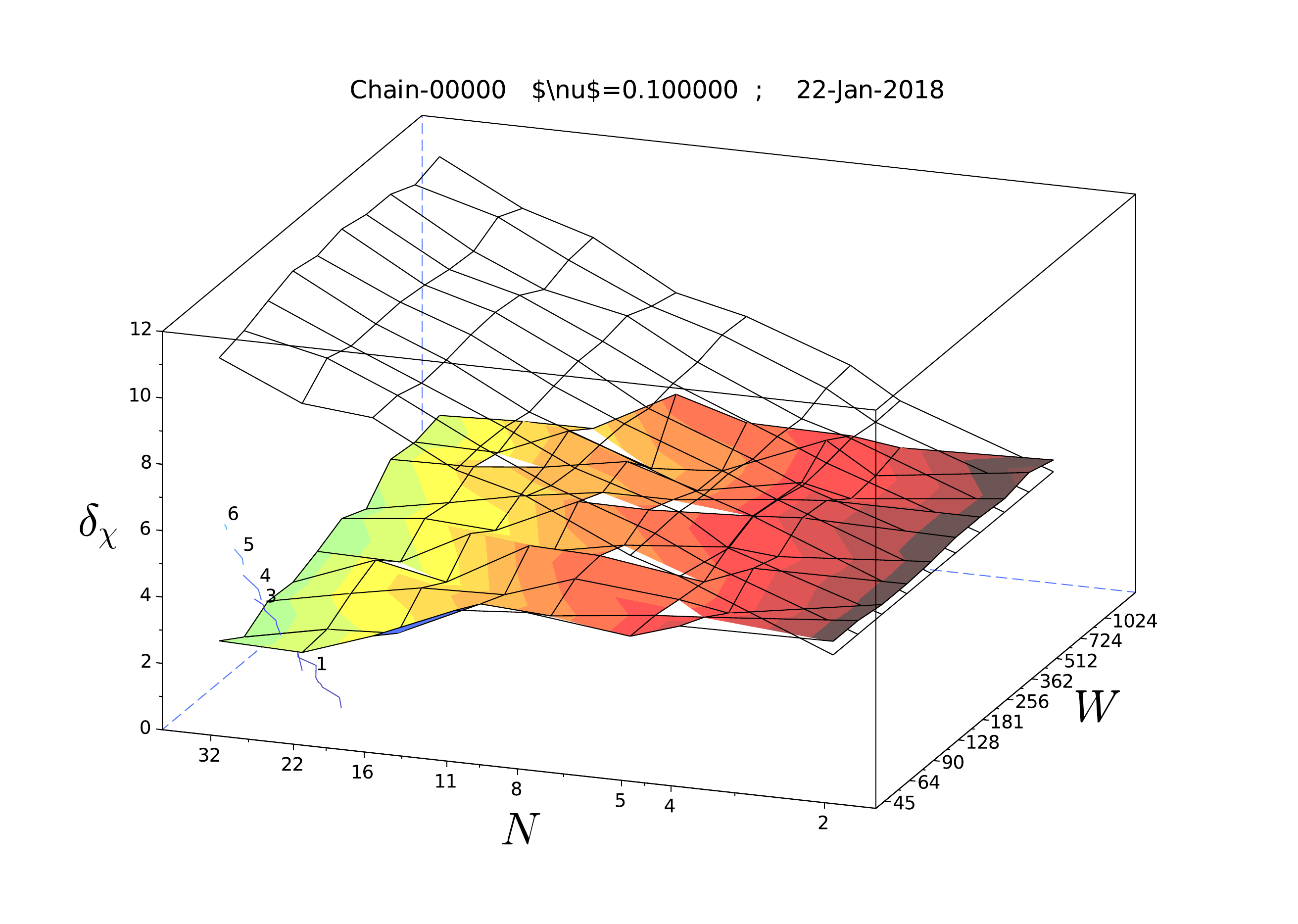}
\caption{Scaling of innovation in
 $\Dict{C}^{\DcSize}_{\SySize,\pFfork}$ chain dictionaries, 
 with symbol number $\SySize$ varying on a log scale from $2$ to $32$, 
 and dictionary size $\DcSize$ on a log scale from $45$ to $1024$. \\
\STANDARDCAP
The floating contours that appear indicate the number of symbols that 
 (on average)
 are not chosen to appear in the dictionaries.
}
\label{fig-chain-scaling}
\end{figure}

\begin{figure}[h]
\includegraphics[angle=0,width=\plotthreewidth]{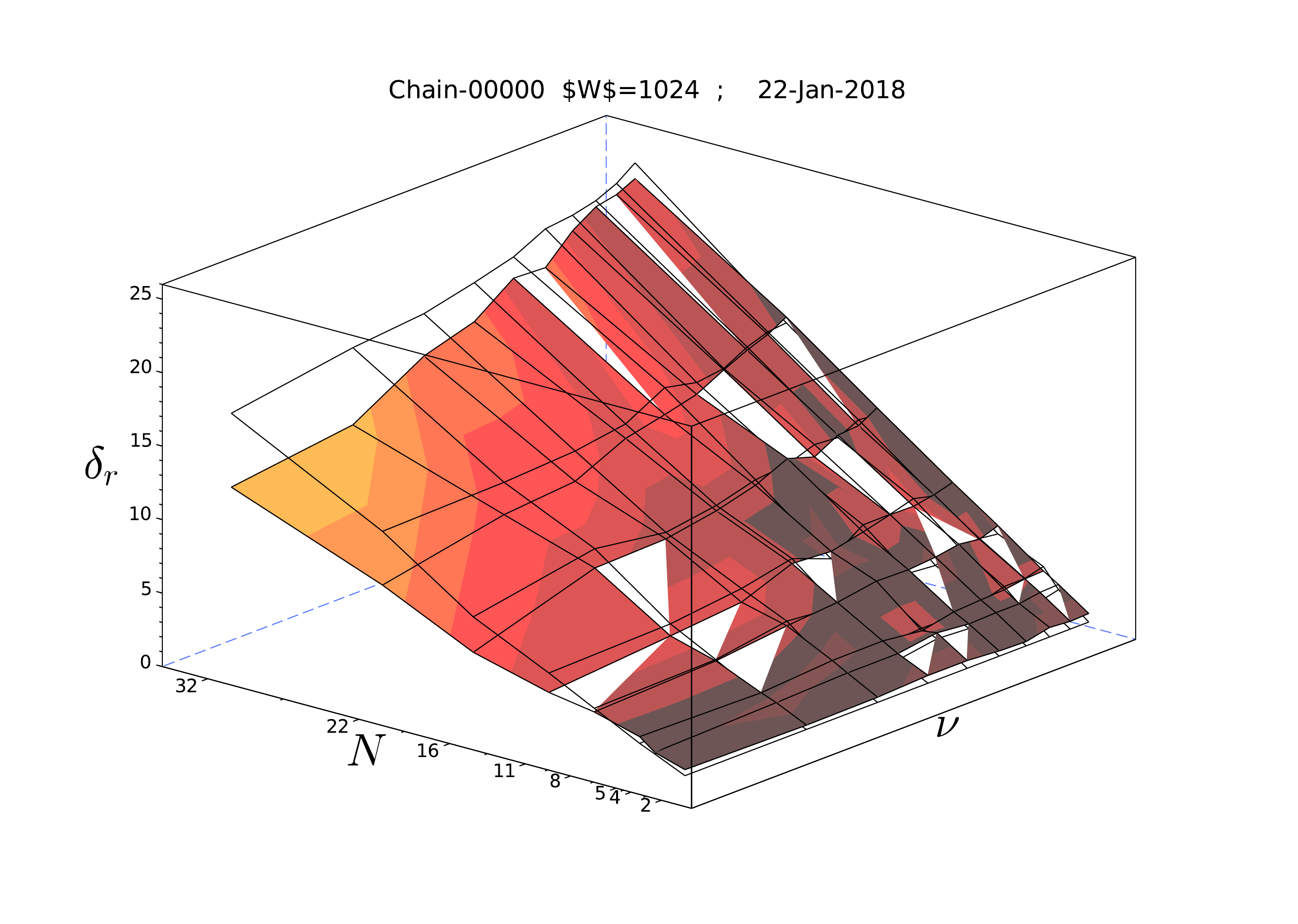}
\includegraphics[angle=0,width=\plotthreewidth]{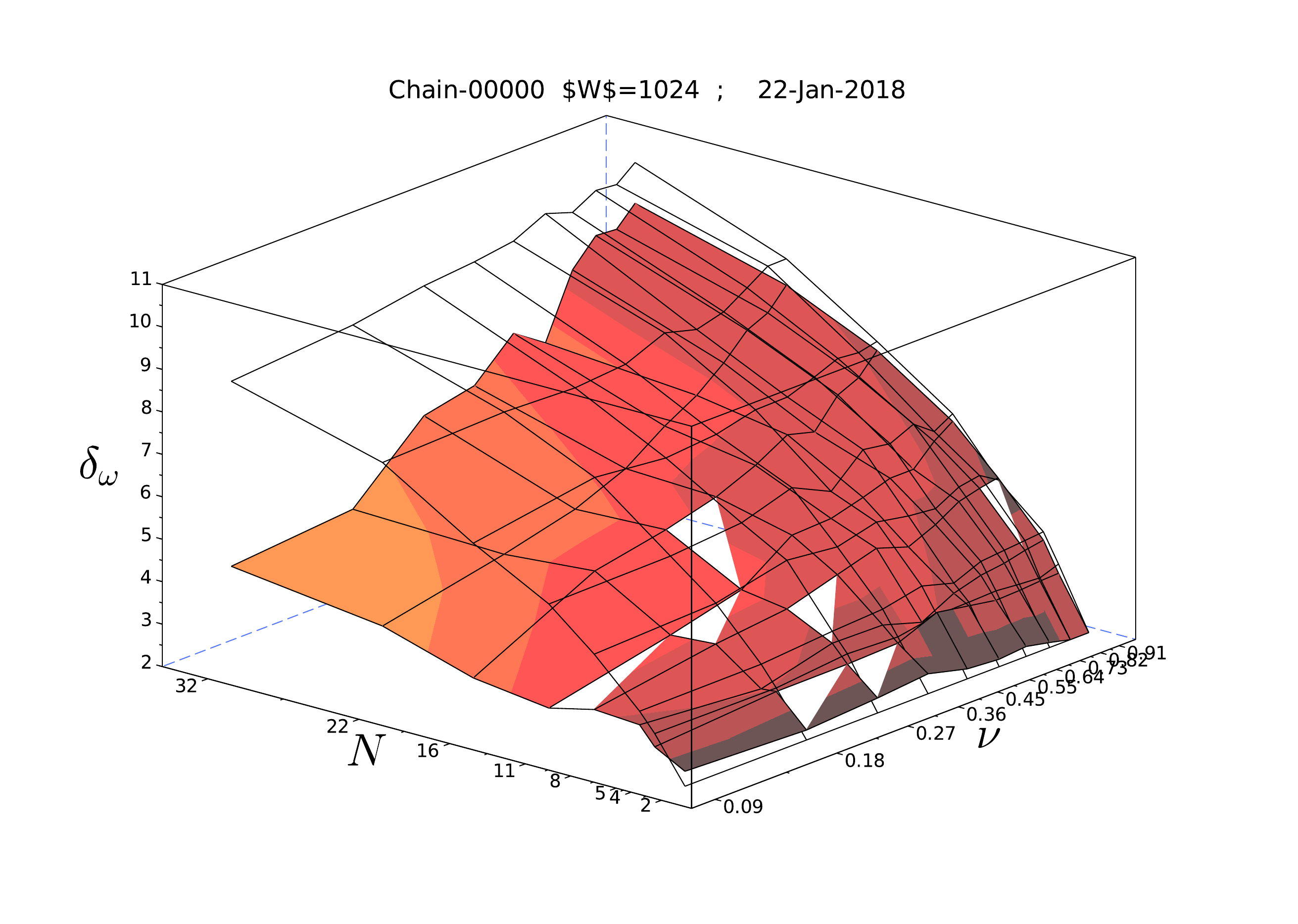}
\includegraphics[angle=0,width=\plotthreewidth]{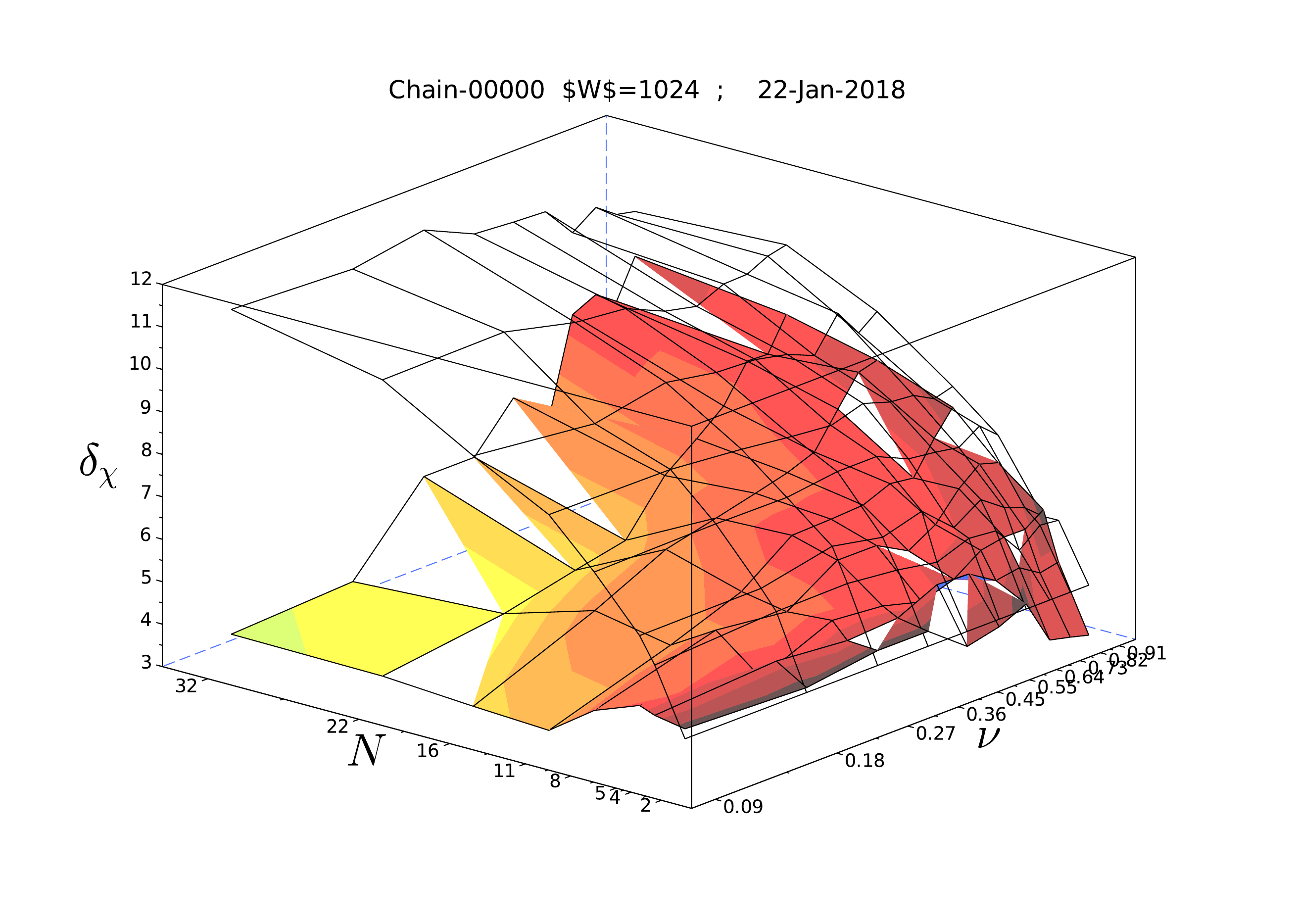}
\caption{Scaling of innovation in
 $\Dict{C}^{1024}_{\SySize,\pFfork}$ chain dictionaries, 
 with symbol number $\SySize$ varying on a log scale from $2$ to $32$, 
 and fork probability $\pFfork$ from $1/11$ to $10/11$.\\
\STANDARDCAP
}
\label{fig-chain-parscaling}
\end{figure}

Fig. \ref{fig-chain-scaling} compares the innovation measures 
 for two symbol discovery possibilities:
 (a) discovery the frequency-order of symbols in the world dictionary, 
 or, 
 (d) discovery in random-order.
The two sets of results here 
 span a range of symbol list and dictionary sizes, 
 but are constrained by a fixed word fork probability $\pFfork=10\%$. 
For low word counts, 
 and large enough symbol lists, 
 some symbols might not be chosen for use in 
 words in the dictionary; 
 the average number of unused symbols 
 is indicated by the floating contours on the figure.

There is no strong trend with dictionary size, 
 but in a new feature not present for the previous dictionary types, 
 discovery order now plays a very strong role:
 discovery in frequency-order shows the expected innovation increase
 with $\SySize$, 
 whereas for random-order discovery the trend is weak (for $\delta_r$), 
 or negligible (for $\delta_\omega, \delta_\chi$).
Once might, 
 for example, 
 even hazard a comparison between the behaviour of F-dictionary behaviour
 and this for the random-order discovery; 
 despite the fact that the frequency-order discovery 
 looks more like that for N- or E-dictionaries!
Indeed, 
 this is again probably a signature of the normalization effect:
 the first (or first few) symbols used in the dictionary generation process 
 will appear in the largest fraction of words.
In frequency order discovery these innovation bursts
 occur early and relatively isolated, 
 in random order they may well appear late and amongst a significant background.
Naturally, 
 such effects stand out more in small dictionaries, 
 where change plays a greater role.
The strong discover-order effect
 is also seen on fig. \ref{fig-chain-parscaling},
 where $\DcSize$ is held fixed at $1024$, 
 and fork probability $\pFfork$ is varied:
 for $\pFfork$,
 the early generated symbols are more dominant, 
 and so randomness in the generation process plays a stronger role.

\subsection{Blinkered: B-dictionaries}

A Blinkered dictionary $\Dict{B}^{\DcSize}_{\SySize,\pFfork}$
 containing $\DcSize$ words is generated
 using a symbol list of length $\SySize$.
It is an incremental process,
 based on a starting point consisting
 of an initial random symbol $\alpha_\circ$, 
 and is somewhat similar to a chain dictionary, 
 but with new symbols only arising due to forks.
Each new word is added iteratively:
 either by
 (i) with probability $1-\pFfork$, 
  one existing word is randomly chosen, 
  a second randomly chosen existing word is appended;
  but is only kept if the result is not already present;
 or 
 (ii) with probability $\pFfork$, 
  another (single) random symbol is chosen to be the new word, 
  but is only kept if the result is not already present.

In this way the fork-fraction $\pFfork$ determines how fast 
 the dictionary accumulates words with different symbols.
Although we should note that the starting symbol has no special role,
 a low $\pFfork$ means that the early-chosen symbols, 
 and in particular $\alpha_\circ$,
 will dominate 
 the symbol frequencies in the resulting dictionary.

\begin{figure}[h]
\includegraphics[angle=0,width=\plotthreewidth]{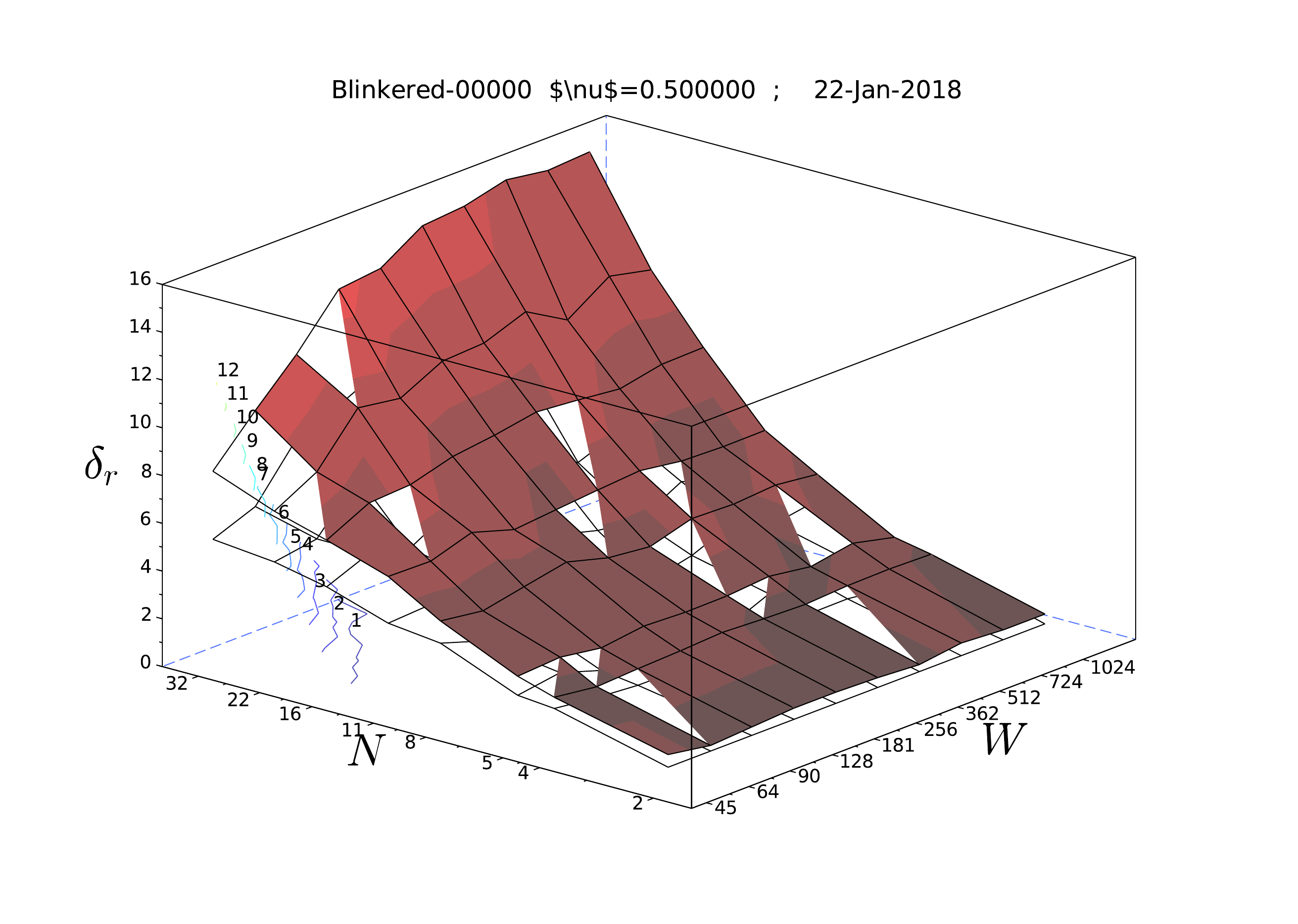}
\includegraphics[angle=0,width=\plotthreewidth]{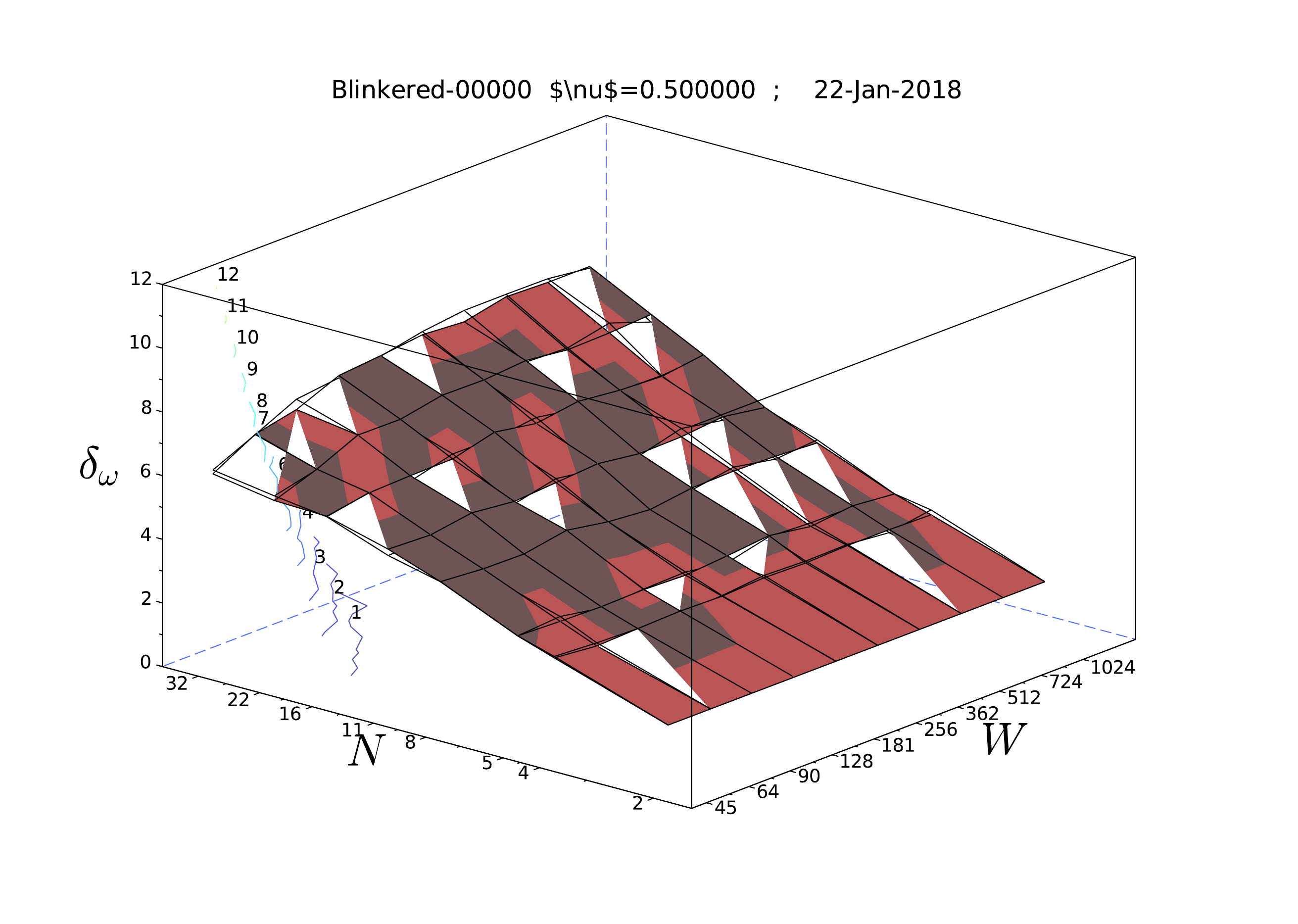}
\includegraphics[angle=0,width=\plotthreewidth]{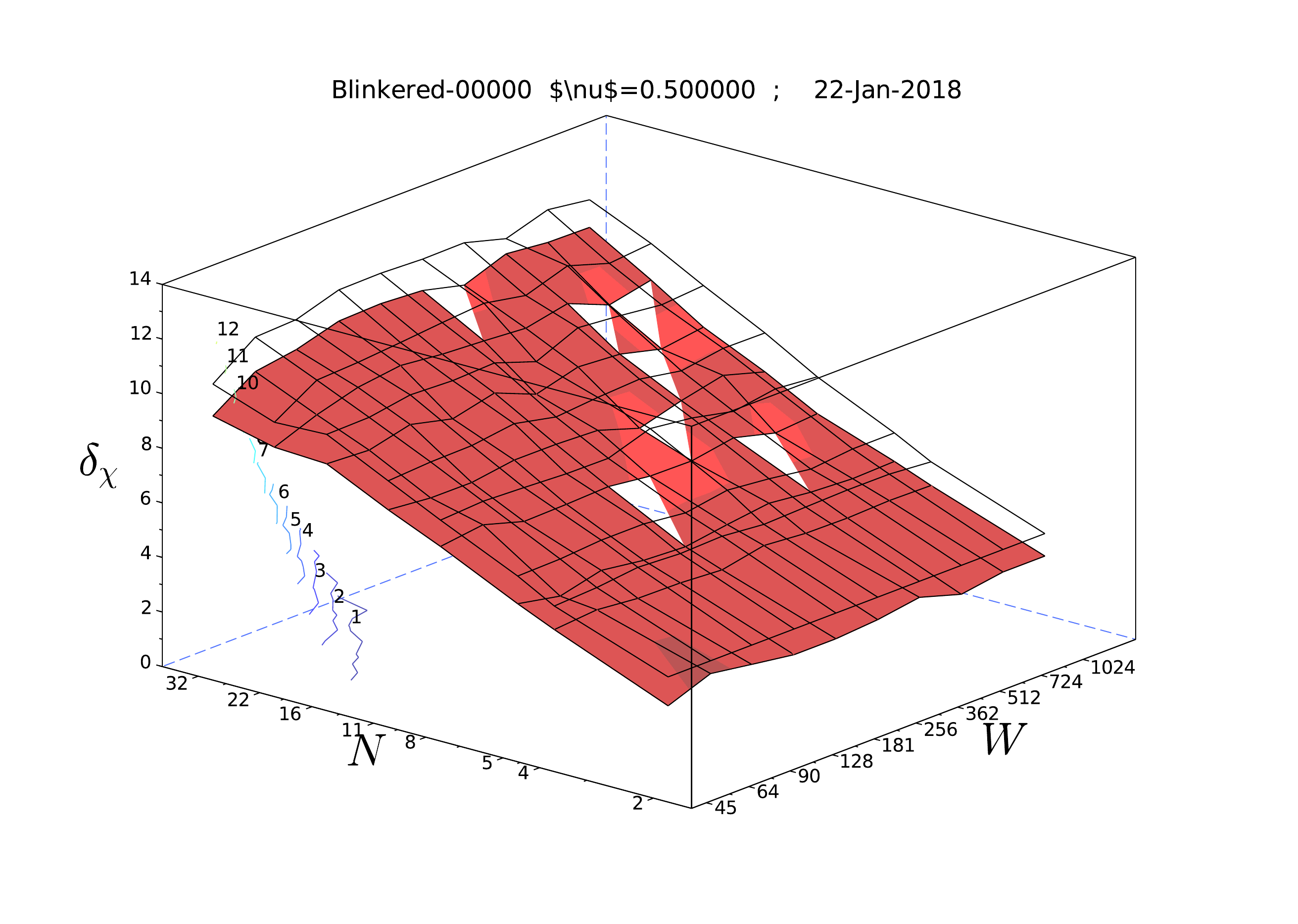}
\caption{Scaling of innovation in
 $\Dict{B}^{\DcSize}_{\SySize,\pFfork}$ blinkered dictionaries, 
 with symbol number $\SySize$ varying on a log scale from $2$ to $32$, 
 and dictionary size $\DcSize$ on a log scale from $45$ to $1024$. \\
\STANDARDCAP
The floating contours that appear indicate the number of symbols that 
 (on average)
 are not chosen to appear in the dictionaries.
}
\label{fig-blinkered-scaling}
\end{figure}

\begin{figure}[h]
\includegraphics[angle=0,width=\plotthreewidth]{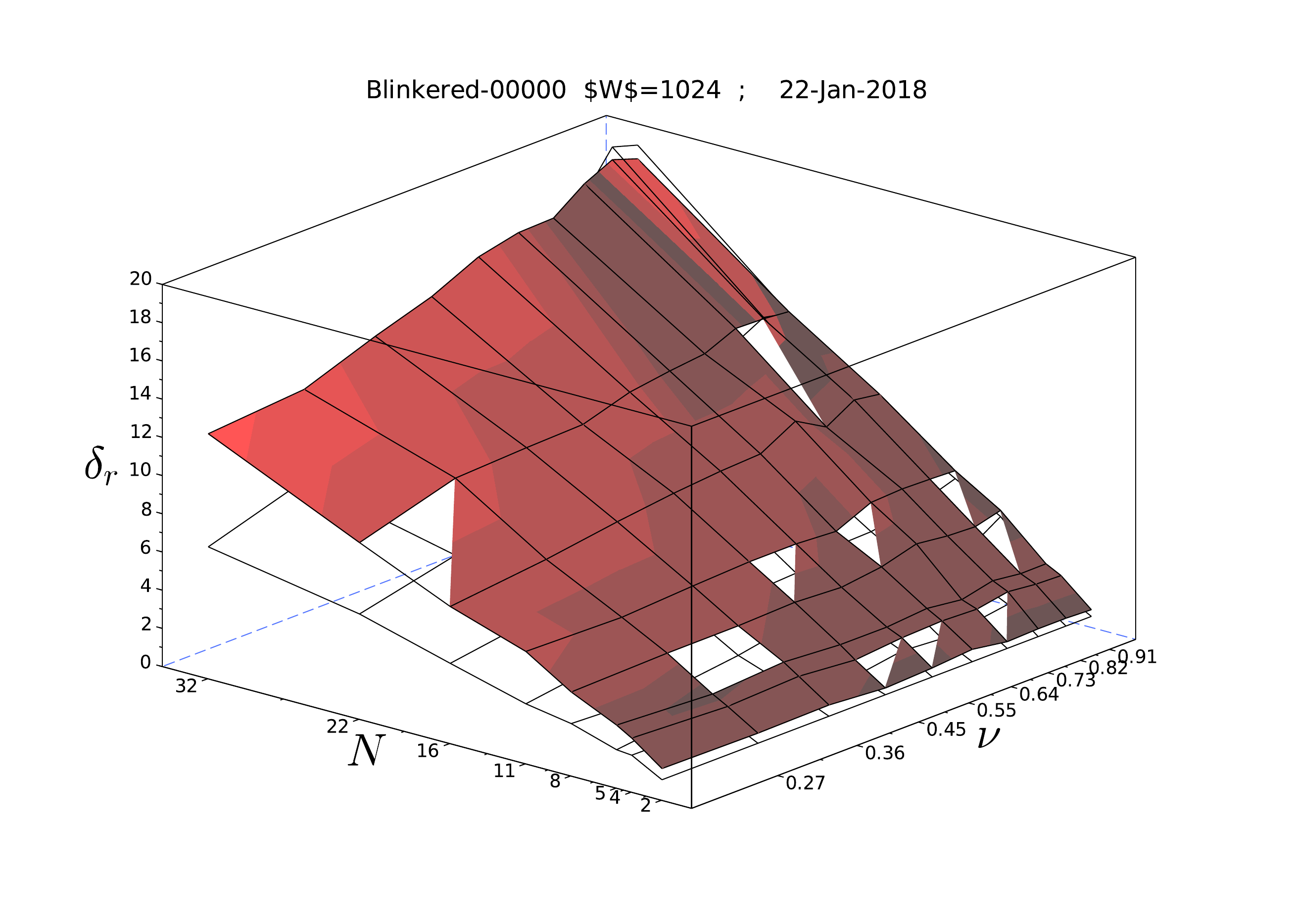}
\includegraphics[angle=0,width=\plotthreewidth]{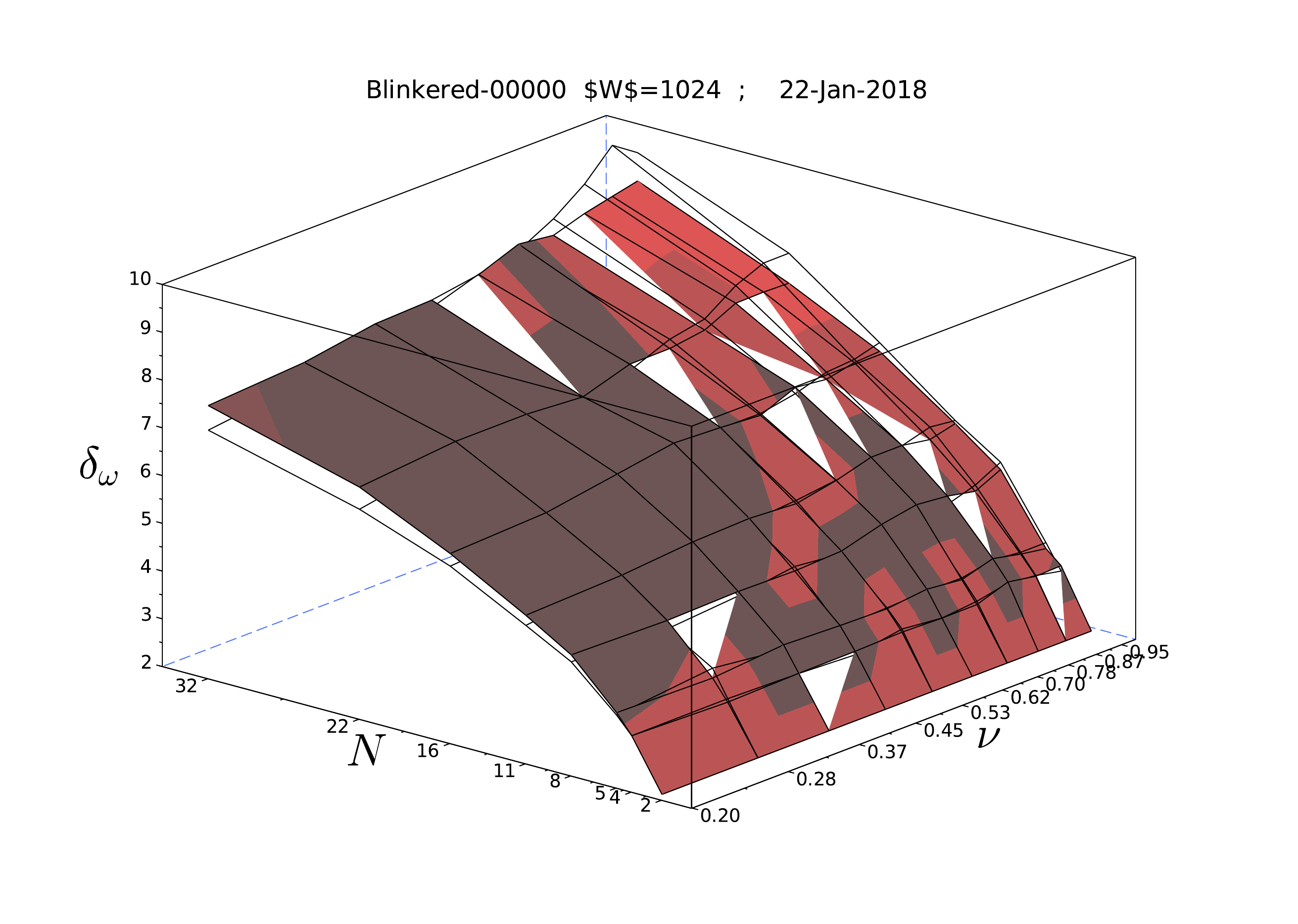}
\includegraphics[angle=0,width=\plotthreewidth]{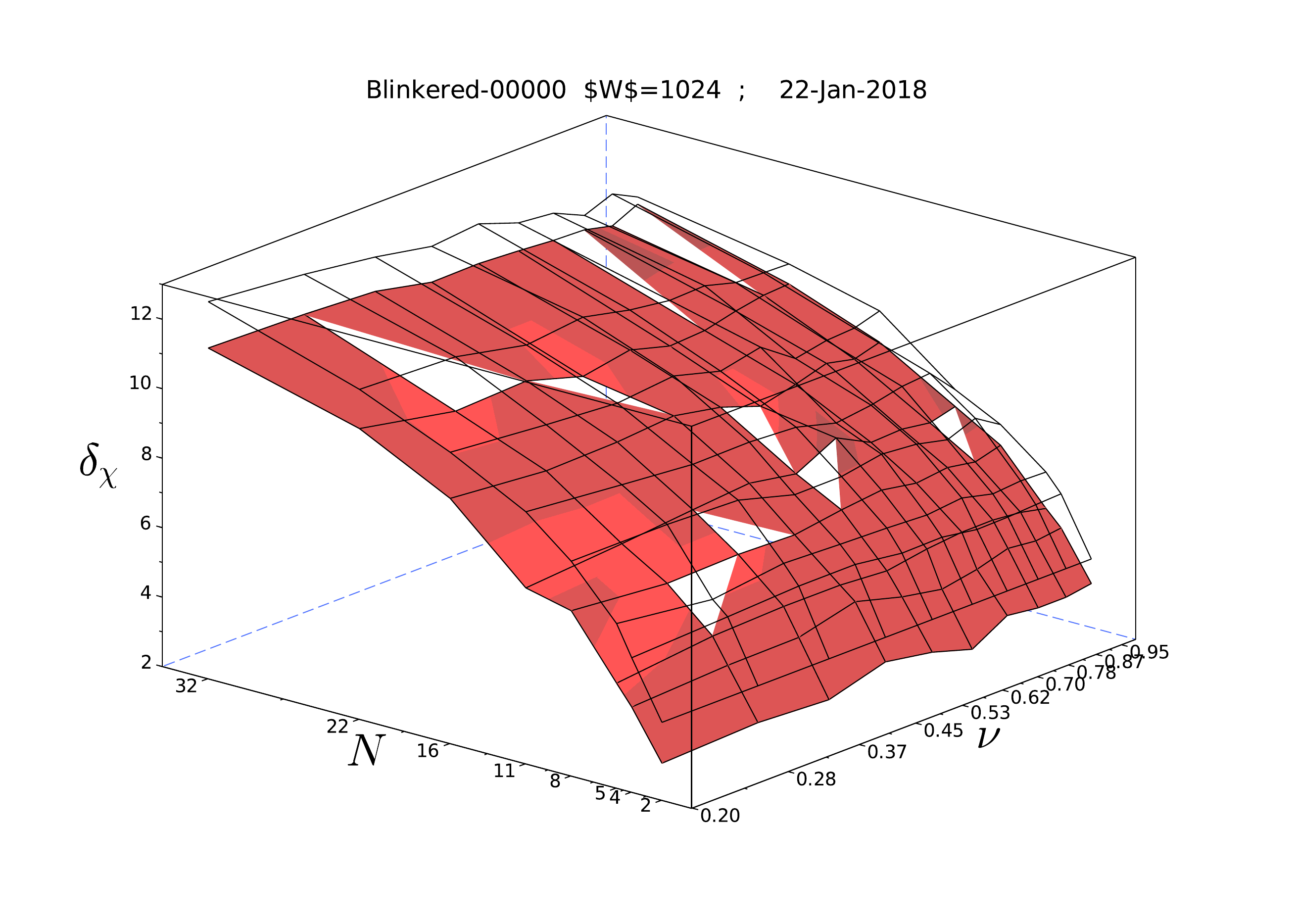}
\caption{Scaling of innovation in
 $\Dict{B}^{\DcSize}_{\SySize,\pFfork}$ blinkered dictionaries, 
 with symbol number $\SySize$ varying on a log scale from $2$ to $32$, 
 and fork probability $\pFfork$ from $1/11$ to $10/11$.\\
\STANDARDCAP
}
\label{fig-blinkered-parscaling}
\end{figure}

Fig. \ref{fig-blinkered-scaling} compares the innovation measures 
 for two symbol discovery possibilities:
 (a) discovery the frequency-order of symbols in the world dictionary, 
 or, 
 (d) discovery in random-order.
The two sets of results here 
 span a range of symbol list and dictionary sizes, 
 but are constrained by a fixed word fork probability $\pFfork=10\%$. 
For low word counts, 
 and large enough symbol lists, 
 some symbols might not be chosen for use in 
 words in the dictionary; 
 the average number of unused symbols 
 is indicated by the floating contours on the figure.
Note that the number of unused symbols tends to be very much higher 
 than for the E- or C- dictionaries, 
 this is because a successful ``fork'' is now the \emph{only} way
 for new symbols to appear in the evolving dictionary.
Nevertheless, 
 despite this distinct unused-symbol behaviour, 
 the  average number of symbols extant in the dictionaries 
 still increases with increasing $\SySize$.

For these B-dictionaries, 
 there is no strong trend with dictionary size, 
 unlike for F- and E-dictionaries; 
 but now higher innovation $\delta_r$ is reported for random discovery order,
 in this is is like the N-dictionary results; 
 but this is not seen for $\delta_\omega$ or $\delta_\chi$.

To address the effects of fork probability,
 in fig. \ref{fig-blinkered-parscaling} 
 I instead hold $\DcSize$ fixed at $1024$, 
 and vary $\pFfork$.
The main feature of note here is that 
 there is a downward trend in reported innovation $\delta_r$ 
 for small $\pFfork$, 
 especially for large symbol lists; 
 in this it is similar to a C-dictionary.

\subsection{Discussion}

An important feature of the results shown here is the role
 of discovery order, 
 and how for some dictionary generation algorithms
 it plays a key role, 
 but for others it does not.
In particular, 
 the significance of discovery order for C-dictionaries
 is rather large, 
 but for B-dictionaries less so --
 this being despite the fact that the generation of B-dictionaries
 is much more contingent on the generation order of symbols, 
 and so we might expect that to influence the effect of discovery order also.
Of course, 
 a strong caveat here is that these are results \emph{averaged} 
 over many dictionaries, 
 and so do not give a clear picture of 
 the experience of any one specific symbol discovery trajectory.
Consequently, 
 I address this point in particular in the next section.



\section{Discovery}
\label{S-Discovery}

\def\stdW{1024}
\def\stdN{32}

In the previous section I looked at the  measures if innovation, 
 as averaged over the entire discovery process from beginning to end.
However, 
 as stated in the introduction, 
 here I want to investigate innovation as a dynamic process
 evolving into an unknown future.
So, 
 with the aggregated results of the previous section as a backdrop, 
 I will now look at some specific examples, 
 based on a given dictionary but different discovery sequences. 
This will usually be 
 a frequency-order discovery of new symbols, 
 as well as two different random-order sequences.
There are of course other options -- 
 the pessimistic might try reverse-frequency-order, 
 where little-used symbols are discovered first;
 or a frequency-weighted random discovery process.

The data plotted are as follows:

\begin{itemize}

\item[(a)]
 Change in average symbol frequencies:
 the average and standard deviation
 of the symbol frequencies at each discovery event are calculated, 
 then used to display the \emph{change} in these between 
 such knowledge at this event and the previous event.
This is plotted on a log scale, 
 with 1 added to the absolute value of the change,
 so that any visible peaks are indicative of a significant innovation event
 where many new words become available.
A higher curve, 
 being the average value plus one standard deviation (in the mean)
 is also shown. 

\item[(b)]
 Symbol entropy, 
 calculated as $E_n = -\sum p_{i,n} \log_2 p_{i,n}$,
 where $p_{i,n}$ is the probability of symbol $i$ being in a word knowable
 at the $n$-th symbol discovery.

\item[(c)]
 Symbol rankings:
However, 
 these are not the raw rankings according to the latest information from 
 only this discovery event, 
 but are averaged over the process so far, 
 then re-ranked.
Processing is done according to the default method in R, 
 so that tied rankings are given an average value
 (i.e. a tie at 3 and 4 results in both being ranked 3.5).

\end{itemize}

\subsection{Fixed length words}

Here we discuss the symbol ranking evolution 
 and known word counts for a sample $\Dict{F}^{\stdW}_{\stdN,8}$ F-dictionary.
Since the generation algorithm
 and word length chosen means that it is extremely unlikely 
 that a word consisting of a single symbol will be in the dictionary, 
 we see that many symbol discoveries are needed before 
 \emph{any} words from the dictionary are known.
This delayed onset is a particular characteristic property
 of this kind of fixed-length word dictionary --
 or, 
 indeed,
 of related types of dictionary 
 which contain only finite minimum length words
 of randomly chosen symbols.
Another characteristic is that -- 
 once started --
 the amount of word innovation accelerates
 as more symbols are discovered.
However, 
 note that this acceleration in innovation is reliant on 
 the size of the dictionary being much less 
 than the number of possible words.

The abbreviated symbol entropy curves
 occur because it cannot be calculated
 for sub-dictionaries which contain no words.

\subsubsection{Frequency order discovery ($\Dict{F}$)}

For frequency-order symbol discovery,
 the main point to note is that the discovery process is highly dynamic, 
 with frequent changes in symbol ranking.
However,  
 these many changes average out to provide relatively smooth 
 progress in the resulting word discovery rate.
Since the most frequent symbols
 are in so many words, 
 the start of word innovation is delayed.
In the example shown in fig. \ref{fig-fixed00}, 
 it is delayed until $8+$ symbols are known --
 and there is another lesser holdup until about the 15th discovery.
These bottlenecks are visible not only in the word rankings, 
 but also the lack of changes in average frequency, 
 and by the premature plateau in symbol entropy.

\def\STANDARDCAPD{Upper frame:
 The black line 
 shows the change in average symbol frequency at each new discovery, 
 with the dotted line showing that average plus one standard deviation
 in the mean.
The red dashed line shows the symbol entropy $E_n$, 
 with the red bar on the right 
 indicating the range from zero to $\log_2(\stdN)$.
The green dot-dashed line shows the fraction of words discovered. 
Lower frame: 
 The coloured lines show the symbol frequency rankings,
 as averaged over previous discovery steps, 
 then re-ranked.
Identical ranks are averaged, 
 so that (e.g.) if two symbols are ranked equal third,  
 the average rank is 3.5.
}

\begin{figure}[h]
\includegraphics[angle=-0,width=0.82\columnwidth]{\FIGDIR{dist_Fixed_R0000}}
\caption{Summary of innovations from a sample
 $\Dict{F}^{\stdW}_{\stdN,8}$ fixed F-dictionary, 
 given frequency-order discovery.
\STANDARDCAPD
}
\label{fig-fixed00}
\end{figure}


\subsubsection{Random discovery ($\Dict{F}$)}

In some respects, 
 random discovery provides a rather similar picture to
 the frequency-order symbol discovery --
 after a sudden initial onset,
 there are many ranking changes,
 but a smooth progress in the resulting word innovation.
This is unsurprising, 
 since new symbols were added randomly to the dictionary as it was grown
 and so variations in symbol frequency are generally small, 
 as can be seen from the very small standard deviation.

\begin{figure}[h]
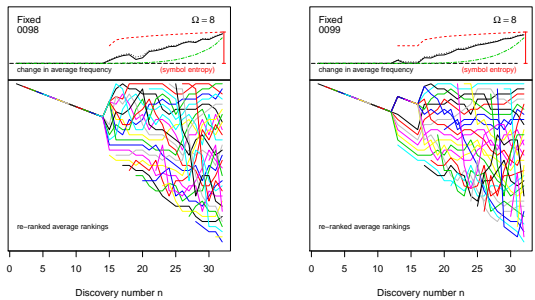

\includegraphics[angle=-0,width=0.45\columnwidth]{\FIGDIR{dist_Fixed_R0098}}
\includegraphics[angle=-0,width=0.45\columnwidth]{\FIGDIR{dist_Fixed_R0099}}
\caption{Summary of innovations from the sample
 $\Dict{F}^{\stdW}_{\stdN,8}$ fixed F-dictionary.
Results for two different randomised symbol discovery orders are shown.
Although individual details differ, 
 the data shown on the upper frames of these plots have a character
 similar to the frequency-order discovery.
In contrast, 
 as can be seen in the lower frames,
 there are very many more changes in symbol ranking,
 How the data is plotted is described in the caption
 to fig. \ref{fig-fixed00}.
}
\label{fig-fixed01}
\end{figure}

By chance, 
 the two random discovery process shown in fig. \ref{fig-fixed01}
 have given distinct outcomes --
 one has the double innovation burst also seen in frequency-order discovery, 
 whereas the other does not.
Nevertheless, 
 both examples have an even more delayed onset than frequency order discovery, 
 being later than the tenth discovery; 
 such behaviour is caused by initial discoveries being of symbols
 that do not appear together.
This behaviour is not unexpected,
 since the number of possible words is very much greater
 than the chosen size of the dictionary.


\subsection{Extensible}

Here we discuss the symbol ranking evolution 
 and known word counts for a sample $\Dict{E}^{\stdW}_{\stdN}$ E-dictionary.
Since the generation algorithm
 means that single symbol words exist, 
 word innovation begins immediately, 
 unlike for the delayed start seen in the F-dictionary case.
Another characteristic
 the amount of word innovation increases much more slowly
 as more symbols are discovered --
 at least as compared to the previous F-dictionary case.

\subsubsection{Frequency order discovery ($\Dict{E}$)}

\begin{figure}[h]
\includegraphics[angle=-0,width=0.82\columnwidth]{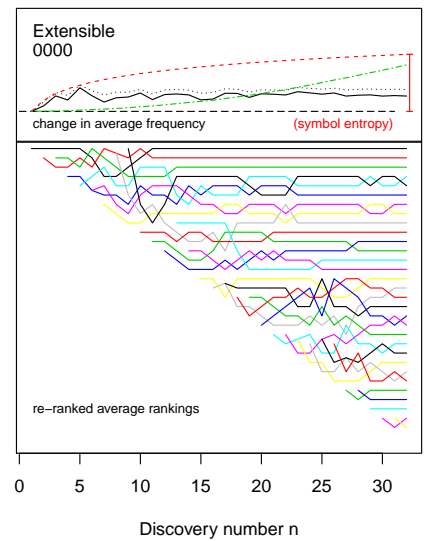}
\caption{Summary of innovations from a sample
 $\Dict{E}^{\stdW}_{\stdN}$ extensible E-dictionary, 
 given frequency-order discovery.
How the data is plotted is described in the caption
 to fig. \ref{fig-fixed00}.
}
\label{fig-extensible00}
\end{figure}

For frequency-order symbol discovery,
 the main point to note is that the discovery process
 shown on fig. \ref{fig-extensible00}
 starts immediately,
 and continues
 with frequent changes in symbol ranking.
However,  
 as for F-dictionaries, 
 these many changes average out to provide gradual 
 progress in the resulting word innovation rate.
While we see that indeed some ``innovation'' occurs, 
 as reflected in the swapping of symbol rankings
 in the lower frame plot, 
 the word increase rate on the upper frame  shows no sudden increases
 in discovered-word counts.
Note also that some of the most frequent symbols do not stay
 as the most highly ranked for the whole discovery process.
Further, 
 the symbol entropy increases smoothly towards its maximum value
 throughout the discovery process.

\subsubsection{Random discovery ($\Dict{E}$)}

\begin{figure}[h]
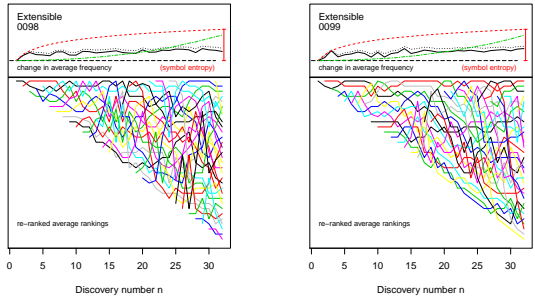

\includegraphics[angle=-0,width=0.45\columnwidth]{\FIGDIR{dist_Extensible_R0098}}
\includegraphics[angle=-0,width=0.45\columnwidth]{\FIGDIR{dist_Extensible_R0099}}
\caption{Different innovation histories from a single sample
 $\Dict{E}^{\stdW}_{\stdN}$ extensible E-dictionary.
Results for two different randomised symbol discovery orders are shown, 
 each with their own distinct experiences of word innovation.
Although individual details differ, 
 the data shown on the upper frames of these plots have a character
 similar to the frequency-order discovery, 
 especially for the symbol entropy.
In contrast, 
 as can be seen in the lower frames,
 there are very many more changes in symbol ranking,
How the data is plotted is described in the caption
 to fig. \ref{fig-fixed00}.
}
\label{fig-extensible01}
\end{figure}

Example of random discovery, 
 as shown on fig. \ref{fig-extensible01}
 provide a rather similar picture to
 the frequency-order symbol discovery. 
This is unsurprising, 
 since new symbols were added randomly to the dictionary as it was grown
 and so variations in symbol frequency are generally small, 
 as can be seen from the small standard deviation.
Nevertheless, 
 we do see that the number of ranking changes is increased
 in comparison to frequency order discovery, 
 although there is little sign of this dynamics
 in the frequency changes or symbol entropy.

\subsection{Chain}

Here we discuss the symbol ranking evolution 
 and known word counts
 for a sample $\Dict{C}^{\stdW}_{\stdN,0.1}$ chain C-dictionary.
Just as for the E-dictionary, 
 since the generation algorithm
 means that single symbol words exist, 
 word innovation can easily begin immediately, 
 unlike for the F-dictionary case.

We see here for the first time a clear difference between 
 the frequency-order and random-order discovery processes.
Both have bursts of innovation, 
 but since the random-order discovery process is unlikely 
 to begin with the symbols used early on in the generation process, 
 we have the potential for a delayed start to word innovation
 while we wait for sufficiently common symbols to be found.
Further, 
 the bursts of innovation are stronger in random discovery, 
 not only adding more structure to changes in symbol rankings, 
 but even slowing or interrupting the increase in symbol entorpy.

\subsubsection{Frequency order discovery ($\Dict{C}$)}

\begin{figure}[h]
\includegraphics[angle=-0,width=0.82\columnwidth]{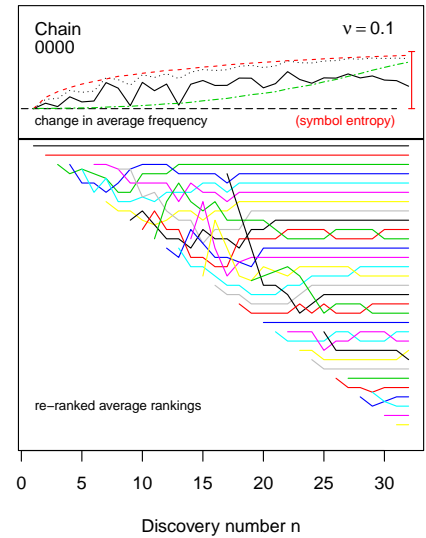}
\caption{Summary of innovations from a sample
 $\Dict{C}^{\stdW}_{\stdN,0.1}$ chain C-dictionary, 
 given frequency-order discovery.
How the data is plotted is described in the caption
 to fig. \ref{fig-fixed00}.
}
\label{fig-chain00}
\end{figure}

Here, 
 new symbols are discovered by interrogators
 in order of their frequency of occurrence 
 in the total dictionary.
While we see on fig. \ref {fig-chain00} that indeed some ``innovation'' occurs, 
 as reflected in the swapping of symbol rankings
 in the lower plot, 
 the upper plot does show some variation --
 perhaps best described as brief bottlenecks --
 in the increase in symbol frequencies. 
However, 
 we can see that whilst the symbol rankings remain more stable
 here that for either the F- or E- dictionaries, 
 the changes in discovered-word counts are more variable.

\subsubsection{Random discovery ($\Dict{C}$)}

\begin{figure}[h]
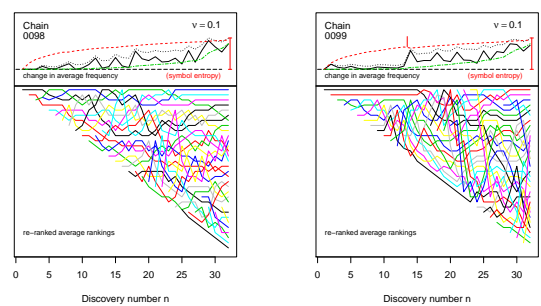

\includegraphics[angle=-0,width=0.45\columnwidth]{\FIGDIR{dist_Chain_R0098}}
\includegraphics[angle=-0,width=0.45\columnwidth]{\FIGDIR{dist_Chain_R0099}}
\caption{Different innovation histories from a single sample
 $\Dict{C}^{\stdW}_{\stdN,0.1}$ chain C-dictionary, 
 the same one as for fig. \ref{fig-chain00} above.
Results for two different random symbol discovery orders are shown, 
 each with their own distinct experiences of word innovation. 
The data shown on the upper frames of these plots have a much more jagged character
 than to the frequency-order discovery, 
 and on the (right) 0099 plot the first strong innovation burst
 is even associated with a small drop in symbol entropy
 (as marked by a vertical red line).
In common with result from previous dictionary types, 
 there are very many more changes in symbol ranking,
 as compared with frequency-order discovery.
How the data is plotted is described in the caption
 to fig. \ref{fig-fixed00}.
}
\label{fig-chain01}
\end{figure}

Unlike the frequency-order discovery for these C-dictionaries, 
 and for the other F- and E-dictionaries,
 here in fig. \ref{fig-chain01} 
 we do see sudden bursts of word innovation, 
 with strong variation in changes in symbol frequency.
Further, 
 although the symbol entropy does not exhibit the step-like change
 that can be seen for F-dictionary results, 
 it is possible that discovery of a new symbol ($\alpha_\star$)
 permits so many new words to be found 
 (i.e. colloquially ``unleashes a wave of innovation'')
 that its ubiquity causes a \emph{drop} in the symbol entropy, 
 as is marked by the vertical bar on fig. \ref{fig-chain01}(right).
This phenomenon is prefigured by a significant period where
 new symbol discoveries have little effect
 on the symbol frequencies --
 presumably these at-the-time apparently uninteresting symbols
 mostly appear in words that also need the then-unknown $\alpha_\star$.



\subsection{Blinkered}

Here we discuss the symbol ranking evolution 
 and known word counts
 for a sample $\Dict{B}^{\stdW}_{\stdN,0.2}$ blinkered B-dictionary.
Not unexpectedly, 
 if we consider the dictionary generation process, 
 the distinction between 
 frequency-order and random-order 
 discovery is by far the clearest here.

When
 new symbols are discovered by interrogators
 in frequency-order, 
 innovation increases rapidly at the start but then 
 flattens out.
In contrast, 
 even with eight possible random-order discovery processes to choose from, 
 there seems to be no commonality between the histories.
There may or may not be a delayed start to innovation, 
 there can be isolated bursts of innovation followed by stagnation, 
 there can be multiple nearby bursts, 
 and these might or might not take place on a background level of innovation.

Lastly, 
 another new feature seen in B-dictionaries
 is that innovation bursts
 can cause the symbol entropy $E_n$
 to under go a small decrease, 
 which briefly counteracts its usual steady increase.
For all other dictionaries, 
 the increase in $E_n$ was monotonic.


\subsubsection{Frequency order discovery ($\Dict{B}$)}

\begin{figure}
\includegraphics[angle=-0,width=0.82\columnwidth]{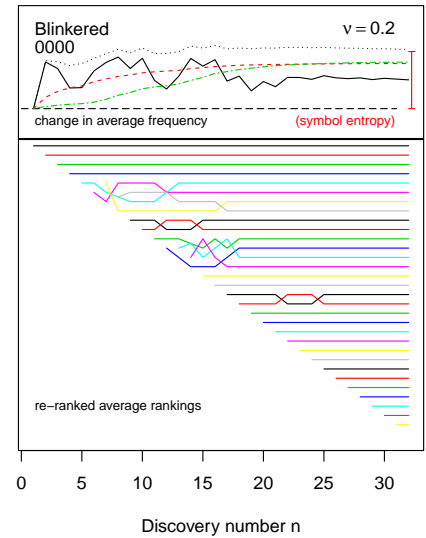}
\caption{Summary of innovations from a sample
 $\Dict{B}^{\stdW}_{\stdN,0.2}$ blinkered B-dictionary, 
 given frequency-order discovery.
How the data is plotted is described in the caption
 to fig. \ref{fig-fixed00}.
}
\label{fig-blinkered00}
\end{figure}

We can see in fig. \ref{fig-blinkered00} a very distinct behaviour --
 most notably, 
 very few changes in ranking.
This is to be expected, 
 given that the inclusion of new symbols was entirely controlled
 by the fork probability, 
 and the discovery order.
A consequence of this is that 
 the change in discovered-word counts tails off significantly
 as the last few -- 
 and rare --
 symbols are discovered.
In fact, 
 because of the way these B-dictionaries are created, 
 frequency-order discovery will match quite closely 
 to the order of appearance of new symbols as the dictionary was generated --
 especially for the beginning of the discovery process.

\subsubsection{Random discovery ($\Dict{B}$)}

\begin{figure}
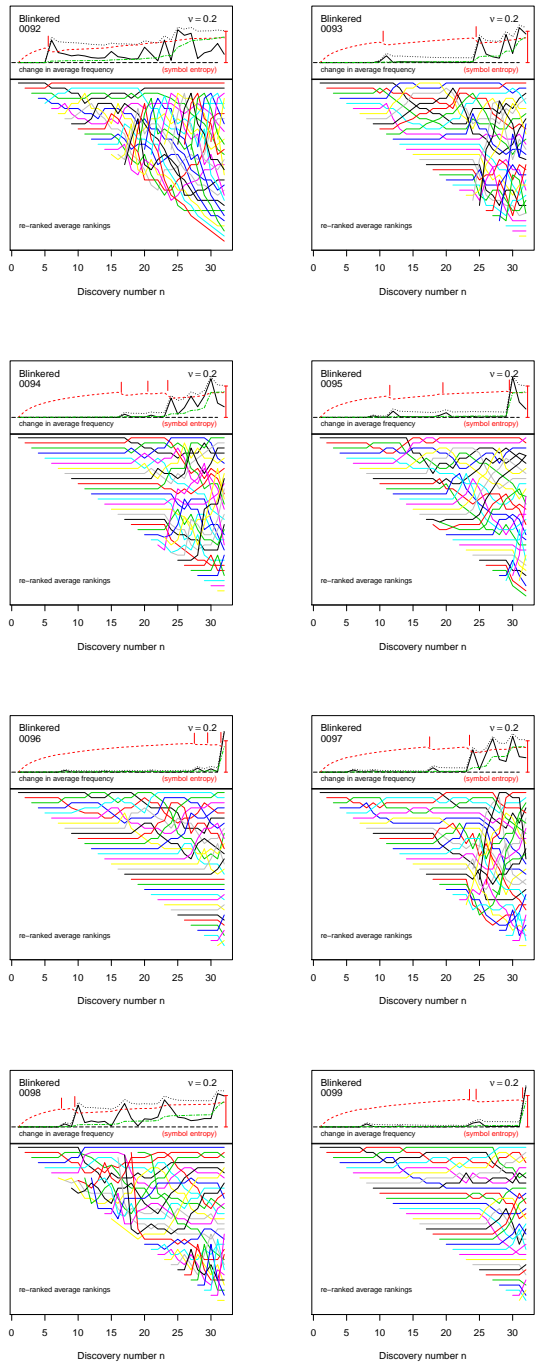

\includegraphics[angle=-0,width=0.45\columnwidth]{\FIGDIR{dist_Blinkered_R0092}}
\includegraphics[angle=-0,width=0.45\columnwidth]{\FIGDIR{dist_Blinkered_R0093}}\\
\includegraphics[angle=-0,width=0.45\columnwidth]{\FIGDIR{dist_Blinkered_R0094}}
\includegraphics[angle=-0,width=0.45\columnwidth]{\FIGDIR{dist_Blinkered_R0095}}\\
\includegraphics[angle=-0,width=0.45\columnwidth]{\FIGDIR{dist_Blinkered_R0096}}
\includegraphics[angle=-0,width=0.45\columnwidth]{\FIGDIR{dist_Blinkered_R0097}}\\
\includegraphics[angle=-0,width=0.45\columnwidth]{\FIGDIR{dist_Blinkered_R0098}}
\includegraphics[angle=-0,width=0.45\columnwidth]{\FIGDIR{dist_Blinkered_R0099}}
\caption{Different innovation histories from a single sample
 $\Dict{B}^{\stdW}_{\stdN,0.2}$  blinkered B-dictionary, 
 the same one as for fig. \ref{fig-blinkered00} above.
Results for eight different random symbol discovery orders are shown, 
 each with it's own distinctive pattern of average frequency changes
 and symbol ranking shifts.
Further, 
 drops in symbol entropy here are both more common than for other dictionaries, 
 and can have larger values.
How the data is plotted is described in the caption
 to fig. \ref{fig-fixed00}.
}
\label{fig-blinkered01}
\end{figure}

Unlike the frequency-order discovery, 
 here we again see sudden bursts of word innovation, 
 as for the random-discovery C-dictionaries, 
 and are shown on fig. \ref{fig-blinkered01}.
However, 
 as the ranking swaps occur, 
 they do so in a more structured way,
 and many new words become available.
These bursts of word innovation will largely be due to the 
 discovery of symbols used early in the dictionary generation process, 
 which we would expect to appear in a large fraction of possible words.
Indeed, 
 the dictionary generation process
 is such that we tend to see rather more of the 
 ``wave of innovation'' induce symbol entropy drops
 in the discovery history than for C-dictionaries.
Again,
 each is generally prefigured by a period where
 new symbol discoveries seem to have little effect, 
 even though in fact each of those apparently unremarkable discoveries 
 act to prime our knowledge base for the burst of innovation made possible
 for when the key symbol is discovered.

\subsection{Discussion}

One point to note is that when comparing different measures
 of the discovery process, 
 we that what might look like innovation under one measure 
 may not necessarily seem like it under another.
In the work of Fink et al  \cite{Fink-RPF-2016arxiv}
 it was changes in rank that was emphasised as a signature of innovation, 
 but it is arguable that other measures --
 perhaps peaks in the change in symbol frequencies, 
 or the brief dips in the symbol entropy provide a better guide.
To determine which is appropriate in any given case
 needs some thought about what we mean by innovation.
This may, 
 of course, 
 be influenced by the underlying dictionary generation model --
 e.g. in F-dictionaries, 
 each of any step increases in the symbol entropy
 looks like a promising signature ...
 but in B-dictionaries, 
 it is the brief \emph{dips} that appear to be the key.
Still, 
 one advantage of symbol entropy is that either signature
 is either present or not --
 whereas the perhaps more robust indicator, 
 i.e. large changes in symbol frequencies, 
 requires some ad hoc definition of the threshold 
 that needs to be passed to be called a ``burst of innovation''.

A feature of the discovery process which may help 
 determine the appropriate measure is the discovery order.
Here I have used frequency-order or random-order, 
 which has been shown to strongly influence
 some proposed innovation signatures.
However, 
 real discoveries are unlikely to be made in the (eventual) frequency order, 
 but it seems implausible that they would be truly random either; 
 although this judgement would likely depend on 
 what sort of innovation space was under discussion.
Perhaps with software component ``symbols'',
 frequency-order might be reasonable, 
 if you assume that effort is spent on solving common problems first, 
 so that those components end up being written (``discovered'') -- 
 more or less --
 according to how much they are eventually used.
In contrast, 
 if applied to a model of basic research, 
 some common problems (``symbols'')
 might be so hard to solve that regardless of effort, 
 other much easier problems are discovered first.

%
\section{Conclusion}
\label{S-Conclusion}

We have seen that the effect of symbol discovery
 allows new word innovation
 in all the model dictionary generators used here.
However, 
 the way the dictionary is generated alters the nature of innovation.

The fixed random F-dictionaries
 required many symbols to be researched
 before any words could be found, 
 after which word innovation improved gradually with each new symbol.
Further, 
 although the mechanism of symbol discovery does affect that process
 in individual cases, 
 it is not a strong effect on average or in general.
The extensible E-dictionaries are more forgiving, 
 with their inclusion of short words allowing early word innovation, 
 but otherwise innovation tended to be gradual, 
 and not on average particularly affected by the mechanism of symbol discovery.

In contrast, 
 the chain C-dictionaries can allow the expectation of bursts
 of innovation, 
 especially under the random symbol discovery model.
This is seen on average, 
 in the scaling data of section \ref{S-Dictionaries} 
 and also in the sample data of section \ref{S-Discovery}.
Notably, 
 we can see such innovation bursts when key symbols
 are discovered in fig. \ref{fig-chain01}.
This characteristic behaviour is cause by its generation process,
 which --
 especially for low fork probabilities $\pFfork$ -- 
 tends to reuse existing words, 
 until a forking event allows new possibilities to emerge.
The crucial event in the discovery process, 
 and then for the number of subsequently available innovations, 
 is therefore finding the symbol which started the generation algorithm.
 
This phenomenon is even more pronounced in blinkered B-dictionaries,
 which are even more dominated by the first (and other early) symbols.
Here, 
 the number of available innovations remains restricted
 until those particular symbols are discovered.

This paper has attempted to evaluate how the nature of the underlying
 innovation space (here, \emph{dictionary})
 might reveal itself as incremental discoveries
 (i.e. of components or \emph{symbols})
 are made, 
 and new innovations (\emph{words})
 become available.
The variation in results has shown that there is indeed some signature, 
 but whether we could expect it to be visible above the variations
 caused by the ad hoc discovery of new components 
 in our single instance of a discovery process
 is debatable.
In particular there is no suggested 
 quantitative measure one might use to try to 
 infer some properties of the underlying innovation space; 
 although the results -- 
 to an extent -- 
 might provide some small basis for generating some intuition.

To summarize, 
 this work leaves us with two tricky problems to address --
 `Which innovation measure is best?',
 and `What discovery order is most appropriate?'; 
 to which we should also add a third --
 `Given some partial discovery/innovation history, 
  is it possible to infer the nature of the underlying dictionary?'.
An answer to this last question --
 assuming it were possible to find one -- 
 might perhaps enable us to optimize our ongoing discovery effort.
All these three questions however are ones that apply to 
 real world discovery/innovation efforts, 
 and not the stylized ones presented here; 
 nevertheless, 
 this purely synthetic approach had the advantage
 of being fully transparent to investigation.

\section*{Acknowledgements}

I am grateful for the support provided by
 EPSRC (the Alpha-X project EP/N028694/1). 

\section*{Methods}

The dictionary generation,
 analysis,
 and plotting
 was done with scripts written for and run using
 Scilab \cite{Scilab} and R \cite{Rlanguage}.

%
%

\begin{thebibliography}{8}%
\makeatletter
\providecommand \@ifxundefined [1]{%
 \@ifx{#1\undefined}
}%
\providecommand \@ifnum [1]{%
 \ifnum #1\expandafter \@firstoftwo
 \else \expandafter \@secondoftwo
 \fi
}%
\providecommand \@ifx [1]{%
 \ifx #1\expandafter \@firstoftwo
 \else \expandafter \@secondoftwo
 \fi
}%
\providecommand \natexlab [1]{#1}%
\providecommand \enquote  [1]{``#1''}%
\providecommand \bibnamefont  [1]{#1}%
\providecommand \bibfnamefont [1]{#1}%
\providecommand \citenamefont [1]{#1}%
\providecommand \href@noop [0]{\@secondoftwo}%
\providecommand \href [0]{\begingroup \@sanitize@url \@href}%
\providecommand \@href[1]{\@@startlink{#1}\@@href}%
\providecommand \@@href[1]{\endgroup#1\@@endlink}%
\providecommand \@sanitize@url [0]{\catcode `\\12\catcode `\$12\catcode
  `\&12\catcode `\#12\catcode `\^12\catcode `\_12\catcode `\%12\relax}%
\providecommand \@@startlink[1]{}%
\providecommand \@@endlink[0]{}%
\providecommand \url  [0]{\begingroup\@sanitize@url \@url }%
\providecommand \@url [1]{\endgroup\@href {#1}{\urlprefix }}%
\providecommand \urlprefix  [0]{URL }%
\providecommand \Eprint [0]{\href }%
\providecommand \doibase [0]{https://doi.org/}%
\providecommand \selectlanguage [0]{\@gobble}%
\providecommand \bibinfo  [0]{\@secondoftwo}%
\providecommand \bibfield  [0]{\@secondoftwo}%
\providecommand \translation [1]{[#1]}%
\providecommand \BibitemOpen [0]{}%
\providecommand \bibitemStop [0]{}%
\providecommand \bibitemNoStop [0]{.\EOS\space}%
\providecommand \EOS [0]{\spacefactor3000\relax}%
\providecommand \BibitemShut  [1]{\csname bibitem#1\endcsname}%
\let\auto@bib@innerbib\@empty
\bibitem [{\citenamefont {Fink}\ \emph {et~al.}(2017)\citenamefont {Fink},
  \citenamefont {Reeves}, \citenamefont {Palma},\ and\ \citenamefont
  {Farr}}]{Fink-RPF-2016arxiv}%
  \BibitemOpen
  \bibfield  {author} {\bibinfo {author} {\bibfnamefont {T.~M.~A.}\
  \bibnamefont {Fink}}, \bibinfo {author} {\bibfnamefont {M.}~\bibnamefont
  {Reeves}}, \bibinfo {author} {\bibfnamefont {R.}~\bibnamefont {Palma}},\ and\
  \bibinfo {author} {\bibfnamefont {R.~S.}\ \bibnamefont {Farr}},\ }\\
\bibfield
  {title} {\bibinfo {title} {Serendipity and strategy in rapid innovation},\ \\
  }\href {https://doi.org/10.1038/s41467-017-02042-w} {\bibfield  {journal}
  {\bibinfo  {journal} {Nat. Commun.}\ }\textbf {\bibinfo {volume} {8}},\
  \bibinfo {pages} {2002} (\bibinfo {year} {2017})},\ \Eprint
  {https://arxiv.org/abs/1608.01900} {1608.01900} \BibitemShut {NoStop}%
\bibitem [{\citenamefont {Marshall}\ \emph {et~al.}(2017)\citenamefont
  {Marshall}, \citenamefont {Murray},\ and\ \citenamefont
  {Cronin}}]{Marshall-MC-2017rsptam}%
  \BibitemOpen
  \bibfield  {author} {\bibinfo {author} {\bibfnamefont {S.~M.}\ \bibnamefont
  {Marshall}}, \bibinfo {author} {\bibfnamefont {A.~R.~G.}\ \bibnamefont
  {Murray}},\ and\ \bibinfo {author} {\bibfnamefont {L.}~\bibnamefont
  {Cronin}},\ }\\
\bibfield  {title} {\bibinfo {title} {A probabilistic framework
  for identifying biosignatures using pathway complexity},\ }\\
\href
  {https://doi.org/10.1098/rsta.2016.0342} {\bibfield  {journal} {\bibinfo
  {journal} {Phil. Trans. Roy. Soc. A}\ }\textbf {\bibinfo {volume} {375}},\
  \bibinfo {pages} {20160342} (\bibinfo {year} {2017})},\ \Eprint
  {https://arxiv.org/abs/1705.03460} {1705.03460} \BibitemShut {NoStop}%
\bibitem [{\citenamefont {Liu}\ \emph {et~al.}(2021)\citenamefont {Liu},
  \citenamefont {Mathis}, \citenamefont {Bajczyk}, \citenamefont {Marshall},
  \citenamefont {Wilbraham},\ and\ \citenamefont {Cronin}}]{Liu-MBMWC-2021sa}%
  \BibitemOpen
  \bibfield  {author} {\bibinfo {author} {\bibfnamefont {Y.}~\bibnamefont
  {Liu}}, \bibinfo {author} {\bibfnamefont {C.}~\bibnamefont {Mathis}},
  \bibinfo {author} {\bibfnamefont {M.~D.}\ \bibnamefont {Bajczyk}}, \bibinfo
  {author} {\bibfnamefont {S.~M.}\ \bibnamefont {Marshall}}, \bibinfo {author}
  {\bibfnamefont {L.}~\bibnamefont {Wilbraham}},\ and\ \bibinfo {author}
  {\bibfnamefont {L.}~\bibnamefont {Cronin}},\ }\\
\bibfield  {title} {\bibinfo
  {title} {Exploring and mapping chemical space with molecular assembly
  trees},\ }\\
\href {https://doi.org/10.1126/sciadv.abj24} {\bibfield  {journal}
  {\bibinfo  {journal} {Science Advances}\ }\textbf {\bibinfo {volume} {7}},\
  \bibinfo {pages} {eabj2465} (\bibinfo {year} {2021})}\BibitemShut {NoStop}%
\bibitem [{\citenamefont {Sood}\ \emph {et~al.}(2010)\citenamefont {Sood},
  \citenamefont {Mathieu}, \citenamefont {Shreim}, \citenamefont
  {Grassberger},\ and\ \citenamefont {Paczuski}}]{Sood-MSGP-2010prl}%
  \BibitemOpen
  \bibfield  {author} {\bibinfo {author} {\bibfnamefont {V.}~\bibnamefont
  {Sood}}, \bibinfo {author} {\bibfnamefont {M.}~\bibnamefont {Mathieu}},
  \bibinfo {author} {\bibfnamefont {A.}~\bibnamefont {Shreim}}, \bibinfo
  {author} {\bibfnamefont {P.}~\bibnamefont {Grassberger}},\ and\ \bibinfo
  {author} {\bibfnamefont {M.}~\bibnamefont {Paczuski}},\ }\\
\bibfield  {title}
  {\bibinfo {title} {Interacting branching process as a simple model of
  innovation},\ }\\
\href {https://doi.org/10.1103/PhysRevLett.105.178701}
  {\bibfield  {journal} {\bibinfo  {journal} {Phys. Rev. Lett.}\ }\textbf
  {\bibinfo {volume} {105}},\ \bibinfo {pages} {178701} (\bibinfo {year}
  {2010})}\BibitemShut {NoStop}%
\bibitem [{\citenamefont {Weiss}\ \emph {et~al.}(2014)\citenamefont {Weiss},
  \citenamefont {Poncela-Casasnovas}, \citenamefont {Glaser}, \citenamefont
  {Pah}, \citenamefont {Persell}, \citenamefont {Baker}, \citenamefont
  {Wunderink},\ and\ \citenamefont {Amaral}}]{Weiss-PGPPBWN-2014prx}%
  \BibitemOpen
  \bibfield  {author} {\bibinfo {author} {\bibfnamefont {C.~H.}\ \bibnamefont
  {Weiss}}, \bibinfo {author} {\bibfnamefont {J.}~\bibnamefont
  {Poncela-Casasnovas}}, \bibinfo {author} {\bibfnamefont {J.~I.}\ \bibnamefont
  {Glaser}}, \bibinfo {author} {\bibfnamefont {A.~R.}\ \bibnamefont {Pah}},
  \bibinfo {author} {\bibfnamefont {S.~D.}\ \bibnamefont {Persell}}, \bibinfo
  {author} {\bibfnamefont {D.~W.}\ \bibnamefont {Baker}}, \bibinfo {author}
  {\bibfnamefont {R.~G.}\ \bibnamefont {Wunderink}},\ and\ \bibinfo {author}
  {\bibfnamefont {L.~A.~N.}\ \bibnamefont {Amaral}},\ }\\
\bibfield  {title}
  {\bibinfo {title} {Adoption of a high-impact innovation in a homogeneous
  population},\ }\\
\href {https://doi.org/10.1103/PhysRevX.4.041008}
 {\bibfield
  {journal} {\bibinfo  {journal} {Phys. Rev. X}\ }\textbf {\bibinfo {volume}
  {4}},\ \bibinfo {pages} {041008} (\bibinfo {year} {2014})}\BibitemShut
  {NoStop}%
\bibitem [{\citenamefont {McNerney}\ \emph {et~al.}(2011)\citenamefont
  {McNerney}, \citenamefont {Farmer}, \citenamefont {Redner},\ and\
  \citenamefont {Trancik}}]{McNerny-FRT-2011pnas}%
  \BibitemOpen
  \bibfield  {author} {\bibinfo {author} {\bibfnamefont {J.}~\bibnamefont
  {McNerney}}, \bibinfo {author} {\bibfnamefont {J.~D.}\ \bibnamefont
  {Farmer}}, \bibinfo {author} {\bibfnamefont {S.}~\bibnamefont {Redner}},\
  and\ \bibinfo {author} {\bibfnamefont {J.~E.}\ \bibnamefont {Trancik}},\
  }\\
\bibfield  {title} {\bibinfo {title} {Role of design complexity in
  technology improvement},\ }\\
\href {https://doi.org/10.1073/pnas.1017298108}
  {\bibfield  {journal} {\bibinfo  {journal} {PNAS}\ }\textbf {\bibinfo
  {volume} {108}},\ \bibinfo {pages} {9008} (\bibinfo {year}
  {2011})}\BibitemShut {NoStop}%
\bibitem [{Sci()}]{Scilab}%
  \BibitemOpen
  \href {http://www.scilab.org/} {\emph {\bibinfo {title} {Scilab}}},\ \bibinfo
  {organization} {Scilab Consortium},\ \bibinfo {note} {software, Open
  Source}\BibitemShut {NoStop}%
\bibitem [{Rla()}]{Rlanguage}%
  \BibitemOpen
  \href {https://www.r-project.org/} {\emph {\bibinfo {title} {R}}},\ \bibinfo
  {organization} {R Foundation},\ \bibinfo {note} {software, Open
  Source}\BibitemShut {NoStop}%
\end{thebibliography}
%

\end{document}